%% file: main.tex
\pdfoutput=1

\documentclass[10pt,twocolumn,letterpaper]{article}

\usepackage[pagenumbers]{cvpr} %

\input{preamble}

\usepackage{microtype}  %
\newcommand{\ours}{\textit{ViDecide}}

\definecolor{cvprblue}{rgb}{0.21,0.49,0.74}
\usepackage{placeins}  %
\usepackage{float}  %
\usepackage[pagebackref,breaklinks,colorlinks,allcolors=cvprblue]{hyperref}

\def\confName{CVPR}
\def\confYear{2027}

\title{ViDecide: Teaching Video Diffusion Models to Reason and Plan Step by Step}

\author{Kaiwen Zhang$^{1}$ \quad
Yiming Gao$^{2}$ \quad
Zeqi Xiao$^{1}$ \\
Xintao Wang$^{3}$ \quad
Pengfei Wan$^{2}$ \quad
Kun Gai$^{2}$ \quad
Xingang Pan$^{1}$ \\ \\
$^{1}$S-Lab, Nanyang Technological University\quad
$^{2}$KlingAI Research\quad
$^{3}$CUHK
}

\begin{document}
\twocolumn[{%
\renewcommand\twocolumn[1][]{#1}%
\maketitle
\input{sec/0_teaser}
}]
\input{sec/0_abstract}
\input{sec/1_intro}
\input{sec/2_relwork}
\input{sec/3_method}
\input{sec/4_exp}
\input{sec/5_conclus}
{
    \small
    \bibliographystyle{ieeenat_fullname}
    \bibliography{main}
}

\input{sec/X_suppl}

\end{document}

%% file: preamble.tex
\usepackage{booktabs}
\usepackage{multirow}
\usepackage{colortbl}
\definecolor{catrow}{gray}{0.92}
\newcommand{\catrow}[2]{\multicolumn{#1}{>{\columncolor{catrow}}l}{\textit{#2}} \\}
\usepackage[most]{tcolorbox}
\definecolor{oursblue}{HTML}{2A78D6}
\newtcolorbox{takeaway}{colback=oursblue!5, colframe=oursblue!75, boxrule=0.6pt, arc=3pt,
  left=5pt, right=5pt, top=3pt, bottom=3pt, before skip=6pt, after skip=6pt}

\usepackage{iftex}
\ifXeTeX
  \usepackage{fontspec}
\fi

%% file: sec/0_teaser.tex
\begin{center}
\centering
\captionsetup{type=figure}
\vspace{-0.4cm}
\includegraphics[width=0.99\linewidth]{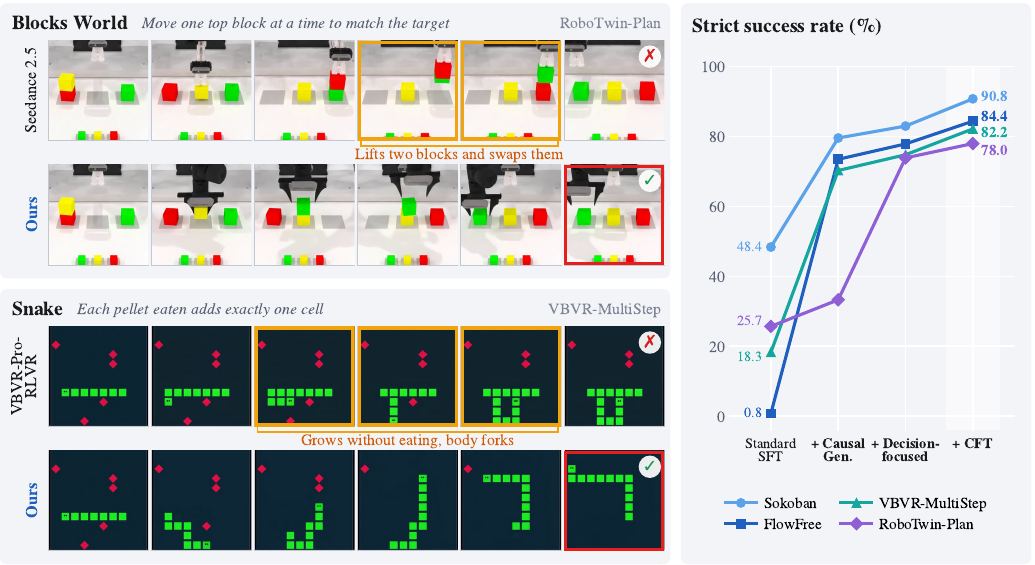}
\vspace{-0.15cm}
\captionof{figure}{\textbf{Left:} Seedance 2.5~\cite{seedance25} hallucinates physically impossible moves to reach the target arrangement, and the reasoning-tuned VBVR-Pro-RLVR~\cite{vbvrpro} breaks the rules of Snake without reaching the goal. Our model solves both puzzles with valid state transitions and stops by itself at an EOS frame (red border). \textbf{Right:} Standard SFT is progressively improved with causal generation, the decision-focused schedule, and Corrective Fine-tuning (CFT). Our designs raise the strict success rate on all multi-step reasoning and planning benchmarks, lifting the average from 23.3\% to 83.8\%.}%
\label{fig:teaser}
\end{center}

%% file: sec/0_abstract.tex
\begin{abstract}
Current leading video generation models can synthesize highly realistic videos, while still frequently violating physical and logical constraints in multi-step reasoning and planning, even after training with large-scale reasoning data or existing diffusion RL methods.
We attribute these failures to the conventional video training paradigm, which treats a solution as a fixed-length clip rather than a chain of decisions.
We present \ours{}, a decision-centric framework that trains video diffusion models to predict state transitions causally and decide when to stop, with a decision-focused training schedule.
We further propose Corrective Fine-tuning (CFT), which corrects failed decisions by mapping their noise to the model's own successful continuations, achieving stable improvement without policy-gradient optimization.
Across abstract games, planning tasks, and a new benchmark of multi-step robot planning, \ours{} achieves an average strict success rate of 83.8\%, far above 23.3\% for standard fine-tuning and 67.8\% for the strongest prior method, and generalizes to larger puzzles and unseen layouts.
Our framework therefore unlocks greater potential for video generation models as reliable reasoners and planners in the visual world.
Project page: \url{https://kevin-thu.github.io/ViDecide}.
\end{abstract}

%% file: sec/1_intro.tex
\section{Introduction}
\label{sec:intro}

Video generation models are emerging as foundation models for visual intelligence, powering world simulation~\cite{genie4,unisim}, embodied planning~\cite{unipi,vlp,cosmospolicy,cosmos3}, and visual reasoning~\cite{videomodelszeroshot,thinkingwithvideo}.
Just as language models reason step by step in words, video models can natively reason and plan in pixel space, unfolding a solution as a sequence of visual states.
This promise has motivated growing efforts to strengthen the reasoning abilities of video models by fine-tuning on large-scale reasoning data~\cite{vbvr,opencof} and reinforcement learning with verifiable rewards~\cite{vbvrpro,videorlvr}.

However, multi-step reasoning and long-horizon planning remain challenging for video diffusion models.
As shown in Fig.~\ref{fig:teaser}, even leading commercial models (\eg, Seedance 2.5~\cite{seedance25}) and models trained on large-scale reasoning data with RL (\eg, VBVR-Pro-RLVR~\cite{vbvrpro}) rarely produce valid solutions.
They often hallucinate state transitions that violate the physical or logical task rules, although the final frame sometimes appears to reach the goal. 
These failures persist even when a video model is fine-tuned on expert solutions of the target tasks, suggesting that more data or rewards alone cannot resolve the problem.

We find that these failures stem from a mismatch between how video diffusion models are trained and what multi-step reasoning requires.
Multi-step reasoning and planning are stateful processes that proceed as a chain of decisions, where each step depends on the previous ones and the reasoning length varies across problems.
Current video diffusion models overlook this structure at three levels.
\emph{First}, they generate all frames of a fixed-length clip jointly.
This ignores the causal order between states and fixes the reasoning length in advance.
\emph{Second}, their training schedule is designed for visual generation, not reasoning.
The model mainly learns to reconstruct visual details at low noise, whereas it must infer state transitions at high noise~\cite{videomodelsreasonearly}.
\emph{Third}, existing diffusion RL methods for video reasoning~\cite{videorlvr,vbvrpro} optimize rewards at the video level.
A single reward thus penalizes the entire video for a few incorrect decisions, which makes credit assignment noisy and provides no explicit correction for them.
These mismatches lead video models to mimic solutions rather than solve problems.

\looseness=-1  %
Our key insight is to treat decisions as the basic units of video reasoning, each corresponding to a visual state transition.
We present \ours{}, a decision-centric framework that trains video diffusion models to reason and plan.
\ours{} generates a reasoning video one segment at a time, each conditioned on all previous states.
To support adaptive termination, we also introduce a visual end-of-sequence (EOS) frame that marks task completion, analogous to the EOS token in language models.
This generation paradigm enforces the causal order of state transitions and adapts the reasoning length to each problem.
Within each segment, reasoning primarily occurs in the early denoising steps, when the next state is not yet visible.
We therefore adopt a decision-focused schedule that shifts flow-matching timesteps toward high noise levels, concentrating training on the decisions themselves.

With these designs, \ours{} can learn valid state transitions from expert videos through supervised training.
At inference, however, the model relies on its own generated states, where a single wrong decision can cause the whole solution to fail.
To correct such failures, ideal supervision should provide explicit corrections for failed decisions and remain within the model's own distribution, making the corrections easier to learn.
Guided by these principles, we propose \textit{Corrective Fine-tuning (CFT)}, a new post-training method for autoregressive video diffusion.
CFT first collects failed rollouts from the current model and traces the decisions that led to the failures.
It then maps the noise of each failed decision to a successful alternative resampled from the same history.
These successful resamples provide explicit, in-distribution corrective supervision, making the model more likely to sample successful decisions.

To demonstrate the effectiveness of \ours{}, we conduct controlled experiments on three benchmarks ranging from abstract puzzles to embodied manipulation.
A solution counts as successful only if it reaches the goal and every state transition follows the task rules.
We first evaluate on the rule-based games of VideoRLVR~\cite{videorlvr} and six planning tasks from VBVR-MultiStep~\cite{vbvr}.
Compared with joint generation, our causal generation with adaptive termination improves the average strict success rate by 48.2 points and outperforms prior video reasoning methods~\cite{hdr,s2pd} on every benchmark.
Compared with training without the timestep shift, the decision-focused schedule improves the average by 37.0 points.
CFT further improves the model on every benchmark, while prior diffusion RL methods~\cite{flowgrpo,diffusionnft} rarely improve it, even when applied to branches from the failed decisions.
These gains also hold on a DiT-B~\cite{dit} trained from scratch on the games.

To extend \ours{} to more complex embodied environments, we further build RoboTwin-Plan, a benchmark of multi-step block-planning tasks for a dual-arm robot in RoboTwin 2.0~\cite{robotwin}.
\ours{} solves 78.0\% of these tasks, compared with 63.9\% for the strongest prior method.
Notably, on these robot videos with rich visual detail, the decision-focused schedule raises the success rate from 33.3\% to 73.9\% over the default schedule of Wan2.2.
Overall, \ours{} raises the strict success rate of standard fine-tuning on all four test sets in Fig.~\ref{fig:teaser}, from 23.3\% to 83.8\% on average.
It also generalizes to larger puzzles and unseen layouts beyond the training range, suggesting that it learns transferable rules rather than only mimicking the training solutions.

In summary, our contributions are as follows:
\begin{itemize}
\item \textbf{A decision-centric framework.} We present \ours{}, which combines causal generation, adaptive termination, and a decision-focused training schedule to enable flexible-length, long-horizon video reasoning and planning.
\item \textbf{Corrective Fine-tuning (CFT).} We propose a new post-training method for autoregressive video diffusion that corrects failed decisions with the model's own successful continuations, achieving stable improvement without policy-gradient optimization.
\item \textbf{State-of-the-art performance.} Extensive experiments on abstract games, planning tasks, and robot manipulation show that \ours{} achieves an average strict success rate of 83.8\%, against 67.8\% for the strongest prior method.
\end{itemize}

%% file: sec/2_relwork.tex
\section{Related Work}
\label{sec:relwork}

\noindent\textbf{Video Generation for Reasoning.}
Recent studies have demonstrated the potential of video generation for visual reasoning \cite{videomodelszeroshot,thinkingwithvideo}, followed by increasingly diverse benchmarks covering spatial and logical tasks \cite{vrbench,mme_cof,vbvr,vbvrpro,vgibench}.
Existing approaches seek to enhance these capabilities through specialized supervision \cite{opencof,vbvr}, reinforcement learning \cite{wanr1,videorlvr,vbvrpro}, and test-time guidance or search \cite{vchain,visualpromptengineering,vlmasteacher,temporalbacktracking}.
Other studies investigate reasoning mechanisms during the diffusion process \cite{demystifyingvideoreasoning,videomodelsreasonearly,scalingvideoreasoning}, or explore structured generation through hierarchical denoising \cite{hdr} and serial-to-parallel diffusion \cite{s2pd}.
Autoregressive video diffusion models~\cite{diffusionforcing,causvid,selfforcing,causalforcing} generate videos causally for long or real-time generation.
Concurrently, ProAR~\cite{proar} also generates reasoning videos autoregressively, with a predetermined video length and additional prospective objectives.
However, all of these methods still treat a reasoning video as a fixed-length clip.
We instead treat a reasoning video as a sequence of decisions and redesign the training and post-training paradigms of video models for reliable multi-step reasoning.

\noindent\textbf{Video-based World Modeling and Planning.}
Video models offer a natural foundation for learning world dynamics and planning through visual generation. 
Early works explore planning for robotic manipulation and autonomous driving through generated visual trajectories \cite{unipi,vlp,epona}. 
The VideoWorld series learns task-relevant knowledge from unlabeled videos through latent dynamics modeling \cite{videoworld,videoworld2}. 
More recent world-action models further adapt large-scale video generation models for embodied control \cite{cosmospolicy,dreamzero,vera,dyna2}. 
Together, these works highlight the potential of video generation to connect world modeling with decision-making.
Our work advances this direction toward reliable task-level planning, with the goal of enabling more capable video-based decision-making systems.

\noindent\textbf{Reinforcement Learning for Diffusion Models.}
Reinforcement learning has become an effective approach to aligning diffusion models with task-specific rewards. 
Policy-based methods adapt PPO/GRPO-style optimization \cite{ppo, grpo} to denoising trajectories \cite{ddpo,flowgrpo,dancegrpo,arcopo}, with recent extensions to video reasoning through verifiable rewards \cite{wanr1,videorlvr,vbvrpro}. 
Such optimization is expensive and hard to stabilize for high-dimensional video generation.
Recent diffusion-specific approaches explore more efficient and stable reward optimization through regression-based objectives \cite{awm,diffusionnft,ram,rvm,diffusionart,worldcompass,creflow}.
However, existing diffusion RL methods mainly assign one reward to a whole sample, although a failed reasoning video often contains only a few wrong moves.
CFT instead uses model-generated successful continuations to provide targeted supervision for the critical decisions.

%% file: sec/3_method.tex
\section{Method}
\label{method}

\begin{figure*}[t]
\centering
\includegraphics[width=\linewidth]{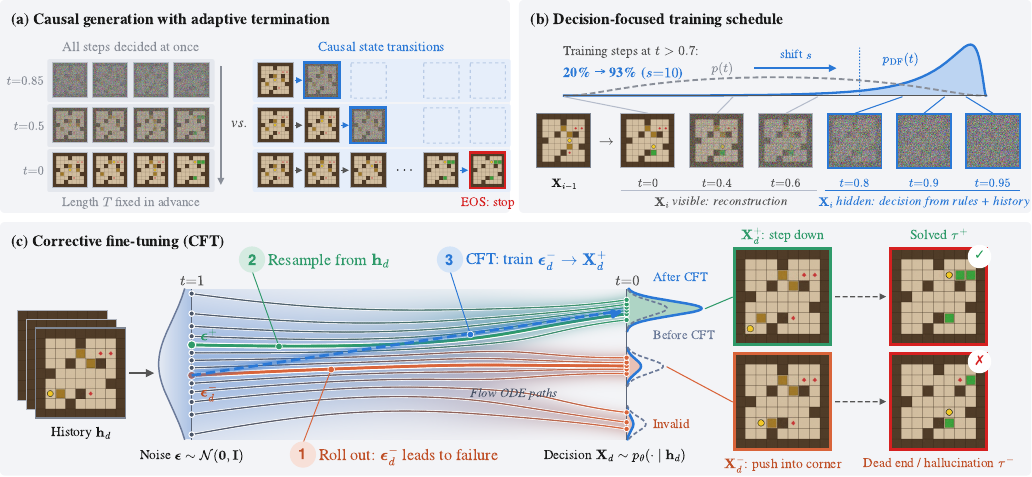}  %
\caption{\textbf{Method overview.} (a) Instead of jointly generating a fixed-length clip, the model generates state transitions causally and ends with a visual EOS frame. (b) The decision-focused schedule moves training to high noise levels, where the next state is hidden and must be inferred from the rules and the history. (c) From the state $\mathbf{h}_d$ before a failed decision, CFT resamples a successful continuation and trains the noise $\boldsymbol{\epsilon}^-_d$ that produced the failed decision to generate the successful decision $\mathbf{X}^+_d$, shifting the decision distribution toward solvable moves.}
\label{fig:pipeline}
\end{figure*}

\subsection{Problem Formulation and Preliminaries}
\label{sec:preliminaries}

\noindent\textbf{Problem Formulation.}
We consider visual reasoning and planning as conditional video generation. 
Given a task description $c$ and an initial observation $\mathbf{o}_0$, the model directly generates a video to solve the task, representing a sequence of visual states $\tau = (\mathbf{o}_0, \mathbf{o}_1, \ldots, \mathbf{o}_T)$.
A successful solution must reach the task goal while maintaining valid state transitions throughout the video.
$R(\tau)\in\{0,1\}$ indicates whether $\tau$ successfully solves the task.
 
\noindent\textbf{Preliminaries on Video Diffusion.}
We adopt flow matching \cite{flowmatching, rectifiedflow} as the training objective for video diffusion models. Let $\mathbf{x}_0$ denote a clean video sample in pixel or latent space, and $\boldsymbol{\epsilon}\sim\mathcal{N}(0,I)$ denote Gaussian noise. At noise level $t\in[0,1]$, the noisy sample is constructed as
\begin{equation}
    \mathbf{x}_t = (1-t)\mathbf{x}_0 + t\boldsymbol{\epsilon}.
\end{equation}
A velocity network $v_\theta$ is trained to predict the corresponding flow field by minimizing
\begin{equation}
    \mathcal{L}_{\mathrm{FM}} =
    \mathbb{E}_{\mathbf{x}_0,\boldsymbol{\epsilon},t\sim p(t)}
    \left[
        \left\|
        v_\theta(\mathbf{x}_t,t;\mathbf{h})
        -(\boldsymbol{\epsilon}-\mathbf{x}_0)
        \right\|_2^2
    \right],
    \label{eq:flow_matching}
\end{equation}
where $\mathbf{h}$ denotes the conditioning information and $p(t)$ is the training noise distribution. During inference, samples are generated by integrating the learned velocity field from noise ($t=1$) to data ($t=0$).

\subsection{Causal Diffusion with Adaptive Termination}
\label{sec:causal_diffusion}

Conventional video diffusion models jointly denoise a fixed-length video, modeling $p_\theta(\mathbf{o}_{1:T}\mid\mathbf{o}_0,c)$ without enforcing the sequential order of decisions. 
Empirically, such models struggle to learn the temporal dependencies imposed by task rules and often produce hallucinated states or invalid transitions.
Recent theoretical analyses further highlight the limitations of parallel diffusion in modeling long chains of dependent events \cite{serialscaling,serialitygap}. 
Moreover, a predefined fixed generation horizon is ill-suited to tasks with varying difficulty and lengths. 
Inspired by the success of autoregressive reasoning in language models, we instead adopt causal video generation with adaptive termination (Fig.~\ref{fig:pipeline}a).

Specifically, we append a visual end-of-sequence (EOS) frame $\mathbf{e}$ to the video trajectory and generate the extended sequence $(\mathbf{o}_1,\ldots,\mathbf{o}_T,\mathbf{e})$
autoregressively. 
At each step $i$, the model generates a temporal segment $\mathbf{X}_i$ spanning up to $K$ consecutive temporal units (pixel or latent frames).
The generation process is factorized as
\begin{equation}
p_\theta(\mathbf{o}_{1:T},\mathbf{e}\mid\mathbf{o}_0,c)
=
\prod_{i=1}^{N}
p_\theta(\mathbf{X}_i\mid\mathbf{X}_{<i},\mathbf{o}_0,c),
\end{equation}
where $N$ is the number of generation steps. 
Each step uses conditional diffusion, trained with standard teacher forcing as in autoregressive language modeling.
This ensures each new segment is generated based on preceding visual states, aligning generation with the causal order of decisions. 
We primarily use $K=1$ for rule-based game reasoning, while larger $K$ suits tasks where each state may last longer, such as embodied manipulation.

Crucially, combining causal generation with a learned visual EOS allows the model to determine when to stop, rather than being constrained by a fixed generation horizon. 
Analogous to the EOS token in language models, the visual EOS is generated after task completion and learned directly through the video generation objective, without a separate termination predictor. 
At inference, generation stops upon detecting this signal, allowing the reasoning trajectory to adapt its length to the task.

\subsection{Decision-Focused Training Schedule}
\label{sec:decision_focused}

While causal generation follows the order of sequential decisions, standard flow-matching training does not distinguish between learning task-relevant state transitions and refining visual details.
At low noise levels, much of the target appearance is already visible in the noisy input, allowing the model to focus on local reconstruction. 
At higher noise levels, however, the target is largely obscured, requiring the model to actually infer what should happen next.

Motivated by this observation, we adopt a decision-focused training schedule that biases flow-matching timestep sampling toward higher noise levels (Fig.~\ref{fig:pipeline}b). 
Specifically, we replace the standard timestep distribution $p(t)$ with a shifted distribution $p_{\mathrm{DF}}(t)$ that assigns greater probability to timesteps closer to $t=1$:
\begin{equation}
t=\frac{s\,t'}{1+(s-1)\,t'},\qquad t'\sim p(t),\; s>1,
\label{eq:shift}
\end{equation}
or equivalently $\operatorname{logit} t=\operatorname{logit} t'+\log s$, where a larger shift $s$ moves more training mass to high noise levels.
The training objective becomes
\begin{equation}
\mathcal{L}_{\mathrm{DF}}
=
\mathbb{E}_{i,t_i,\boldsymbol{\epsilon}_i}
\big[\big\|
v_\theta(\tilde{\mathbf{X}}_i,t_i;\mathbf{h}_i)
-
(\boldsymbol{\epsilon}_i-\mathbf{X}_i)
\big\|_2^2\big],
\label{eq:df}
\end{equation}
where $\tilde{\mathbf{X}}_i=(1-t_i)\mathbf{X}_i+t_i\boldsymbol{\epsilon}_i$, the noise level $t_i\sim p_{\mathrm{DF}}(t)$ is sampled independently for each step $i\in\{1,\ldots,N\}$, and $\mathbf{h}_i=(\mathbf{X}_{<i},\mathbf{o}_0,c)$
denotes the clean causal context.

\subsection{Corrective Fine-tuning}
\label{sec:cft}

Through supervised training, the model learns to reason and plan by imitating expert trajectories step by step.
At inference, however, the model conditions on its own generated history, so a single invalid transition or dead-end move can cause the entire solution to fail.
Reinforcement learning with the success signal $R(\tau)$ can address this in principle, but existing diffusion RL methods spread the reward of a failed video over its many valid transitions, and each transition spans a high-dimensional video segment, so the learning signal for the few wrong decisions is both sparse and noisy.
To address this, we propose \textit{Corrective Fine-tuning (CFT)}, which builds on causal generation to directly supervise failed decisions (Fig.~\ref{fig:pipeline}c).

\noindent\textbf{Collecting self-generated corrections.}
Given the current model, we sample video trajectories $\tau\sim p_\theta(\cdot\mid\mathbf{o}_0,c)$ on the training set and collect those with $R(\tau)=0$ as failed rollouts $\tau^-$.
For each failure, we resample continuations from a few steps before the failure detected by the verifier and keep those with $R(\tau)=1$ as $\tau^+$.
The first step at which $\tau^+$ departs from $\tau^-$ is the divergence step $d$.
Under the same causal context $\mathbf{h}_d$, the failed rollout generates $\mathbf{X}^-_d$, while the successful continuation generates $\mathbf{X}^+_d$.
This yields a self-generated correction pair $(\mathbf{X}^-_d,\mathbf{X}^+_d)$ at the divergence step, enabling targeted supervision of the failed decision.

\noindent\textbf{Training objective.}
To correct the failed decision under its original sampling conditions, CFT reuses the initial noise $\boldsymbol{\epsilon}^-_d$ that produced $\mathbf{X}^-_d$ and pairs it with the corrected segment $\mathbf{X}^+_d$.
Specifically, we construct the corrective flow-matching path $\tilde{\mathbf{X}}^{+}_d=(1-t)\mathbf{X}^+_d+t\boldsymbol{\epsilon}^-_d$ and optimize
\begin{equation}
\mathcal{L}_{\mathrm{corr}}
=\mathbb{E}_{t\sim q(t)}
\big[\big\|v_\theta(\tilde{\mathbf{X}}^{+}_d,t;\mathbf{h}_d)-(\boldsymbol{\epsilon}^-_d-\mathbf{X}^+_d)\big\|_2^2\big].
\label{eq:cft_corr}
\end{equation}
For the remaining suffix of $\tau^+$, we apply the decision-focused flow-matching objective of
Sec.~\ref{sec:decision_focused},
\begin{equation}
\mathcal{L}_{\mathrm{suffix}}
=\mathbb{E}_{i>d,\,t_i,\boldsymbol{\epsilon}_i}\big[\big\|v_\theta(\tilde{\mathbf{X}}^+_i,t_i;\mathbf{h}^+_i)-(\boldsymbol{\epsilon}_i-\mathbf{X}^+_i)\big\|_2^2\big].
\label{eq:cft_suffix}
\end{equation}
Steps before $d$ serve only as context and receive no corrective loss.
For $\mathcal{L}_{\mathrm{corr}}$, $q(t)$ places additional mass at $t=1$, directly supervising the model to generate the corrected decision from the very noise that previously led to failure.
To preserve the model's existing capabilities, we further mix expert trajectories into fine-tuning with the decision-focused objective $\mathcal{L}^{\mathrm{exp}}_{\mathrm{DF}}$ (Sec.~\ref{sec:decision_focused}).
The overall CFT objective is
\begin{equation}
\mathcal{L}_{\mathrm{CFT}}
=\mathcal{L}_{\mathrm{corr}}
+\mathcal{L}_{\mathrm{suffix}}
+\lambda\,\mathcal{L}^{\mathrm{exp}}_{\mathrm{DF}},
\label{eq:cft}
\end{equation}
where $\lambda$ balances corrective and expert supervision.
The procedure can be repeated over multiple rounds using newly generated failures and corrections of the updated model.

\noindent\textbf{Discussion.}
CFT differs from existing reward optimization in where and how it assigns credit.
Compared with trajectory-level policy gradients, its supervision is focused on the decisions that led to the failure, and its target is an explicit correction rather than a scalar penalty on the whole video.
Compared with rejection-sampling fine-tuning on successful videos, CFT pairs each correction with the noise that produced the failure.
For a given visual history, different noise realizations induce a distribution over possible decisions. 
By pairing the noise that led to a failed continuation with a successful alternative, CFT explicitly redirects these failure-prone samples toward corrected decisions. 
Collectively, these corrections encourage the decision distribution to shift toward successful outcomes.
Finally, as a regression objective on the model's own trajectories, CFT enables stable improvement and fast convergence without requiring policy-gradient optimization for high-dimensional video generation.
Together, these properties turn sparse task-level feedback into targeted decision-level supervision for long-horizon visual reasoning.

%% file: sec/4_exp.tex
\section{Experiments}
\label{sec:exp}

To systematically analyze the design of video diffusion models for reasoning and planning, we conduct controlled experiments around three questions, corresponding to the three components of \ours{}:
how to structure the generation process (Sec.~\ref{sec:exp_paradigm}),
how to focus training on decisions (Sec.~\ref{sec:exp_schedule}),
and how to learn from its own failures during RL post-training (Sec.~\ref{sec:exp_cft}).
In each part, we compare our design against representative baselines under a shared protocol and explain why it works.
We further test whether the resulting models generalize beyond the training distribution (Sec.~\ref{sec:exp_ood}).

\begin{table*}[t]
\centering
\small
\caption{\textbf{Generation paradigms} (SSR, \%). VBVR-MultiStep and RoboTwin-Plan report the success rate over all tasks; per-task results are in Tabs.~\ref{tab:paradigm_vbvr} and~\ref{tab:paradigm_robotwin}. Zero-shot generators are judged with a relaxed criterion that accepts non-optimal solutions and minor errors after manual review. $^\dagger$Evaluated on stratified random subsets of 20--30 test samples per benchmark due to API cost.}
\label{tab:paradigm}
\begin{tabular*}{\linewidth}{@{\hspace{\tabcolsep}\extracolsep{\fill}}lcccccc}
\toprule
& \multicolumn{4}{c}{Wan2.2-5B fine-tuned} & \multicolumn{2}{c}{DiT-B trained from scratch} \\
\cmidrule(lr){2-5}\cmidrule(lr){6-7}
Method & FlowFree & Sokoban & VBVR-MultiStep & RoboTwin-Plan & FlowFree & Sokoban \\
\midrule
\catrow{7}{Zero-shot video generators}
Seedance 2.5$^\dagger$ & 0.0 & 5.0 & 20.0 & 6.7 & -- & -- \\
Veo 3.1$^\dagger$ & 0.0 & 5.0 & 0.0 & 0.0 & -- & -- \\
Wan2.2-5B & 0.0 & 0.0 & 0.0 & 0.0 & -- & -- \\
VBVR-Wan2.2-A14B & 0.0 & 0.0 & 11.3 & 0.0 & -- & -- \\
VBVR-Pro-Wan2.2-5B-RLVR & 0.0 & 0.0 & 7.7 & 0.0 & -- & -- \\
\catrow{7}{Fixed-length generation}
Bidirectional & 1.1 & 50.8 & 22.3 & 42.6 & 0.0 & 8.8 \\
HDR~\cite{hdr} & 11.6 & 45.9 & 24.0 & 63.9 & 0.4 & 10.3 \\  %
S2PD~\cite{s2pd} & 68.1 & 75.7 & 64.5 & 62.9 & 41.2 & 56.4 \\  %
\catrow{7}{Causal generation with adaptive termination}
Ours ($K{=}8$) & 28.5 & 75.0 & 57.8 & \textbf{73.9} & 5.8 & 18.3 \\
Ours ($K{=}1$) & \textbf{77.9} & \textbf{83.0} & \textbf{74.8} & 70.1 & \textbf{58.6} & \textbf{73.8} \\
\bottomrule
\end{tabular*}
\end{table*}

\begin{figure*}[t]
\centering
\includegraphics[width=\linewidth]{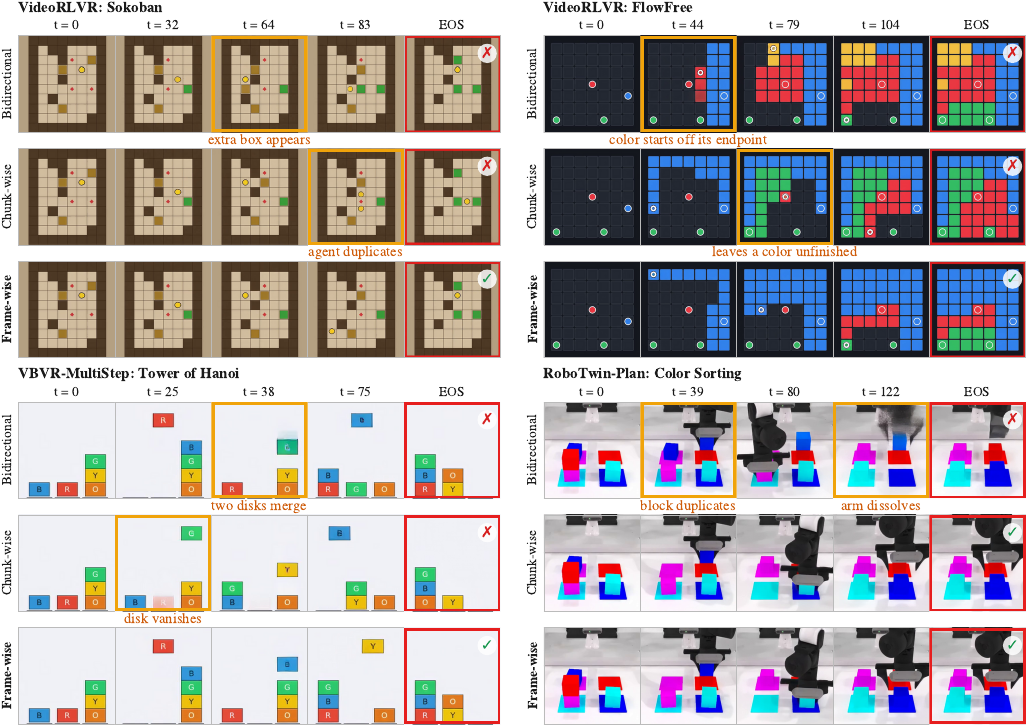}  %
\caption{\textbf{Qualitative comparison of generation paradigms} on one test puzzle per benchmark. The first four columns show the same frame index $t$ in every row, and the last column shows each model's EOS frame (red border). Orange borders mark rule violations.}
\label{fig:qualitative}
\end{figure*}

\subsection{Experimental Setup}
\label{sec:exp_setup}

\noindent\textbf{Benchmarks.}
To demonstrate the effectiveness of our designs, we evaluate on three complementary benchmarks.

\textit{VideoRLVR}~\cite{videorlvr} evaluates abstract reasoning on rule-based games, consisting of Maze, FlowFree, and Sokoban, each with 10K training and 1K test puzzles.
Despite simple visuals, they require long sequences of rule-consistent moves.
We train a specialist model for each game.

\textit{VBVR-MultiStep}, part of the VBVR suite~\cite{vbvr}, is a large-scale collection of procedurally generated tasks for multi-step video reasoning.
We adopt its six planning tasks, namely Sliding Puzzle, Tower of Hanoi, Snake Routing, Sokoban, TSP Reward Collection, and Ordinal Number Sequence, which differ in rules, visual styles, and solution lengths.
A single generalist model is trained on all tasks, with 5K training and 100 test puzzles per task.

\textit{RoboTwin-Plan} extends to multi-step planning tasks in embodied environments.
Using the RoboTwin 2.0~\cite{robotwin} simulator, we build three block-planning tasks for a dual-arm robot: Blocks World, Tower of Hanoi, and Color Sorting, each with two layout variants.
Unlike most logical games, each decision here unfolds over many frames of continuous arm motion.
A single generalist model is trained on 10K episodes and evaluated on 700 test episodes.

\noindent\textbf{Evaluation metric.}
Existing evaluations of these benchmarks give credit to videos that break the rules or compare them with a single reference solution.
We instead report the \emph{Strict Success Rate (SSR)}, which counts a video as solved only if every transition obeys the task rules and the video ends in the goal state~\cite{viper}.
On VBVR-MultiStep, whose prompts ask for the shortest solution in five tasks, solutions with redundant moves are also counted as failures.
For each task, a rule verifier parses the state of every generated frame and replays the trajectory.
Verifier details, including the learned state reader for RoboTwin, are provided in the supplement.

\noindent\textbf{Implementation details.}
Our main experiments are built on Wan2.2-5B~\cite{wan}, which we fully fine-tune on all benchmarks.
On the discrete games in VideoRLVR, we further validate our findings with DiT-B~\cite{dit}, a 132M-parameter pixel-space diffusion transformer trained from scratch.
For Wan2.2-5B, all videos are converted to $384{\times}384$ at 16 fps with their native duration, and every training video ends with four EOS frames that repeat the final state with a red border.
Wan2.2-5B models are trained with AdamW and a cosine learning-rate schedule (peak $10^{-5}$) for 10 epochs on VideoRLVR and RoboTwin-Plan and 5 epochs on VBVR-MultiStep.
To ensure a fair comparison, all variants within each comparison share the same data and training budget, and all models, including the baselines, are trained with $s{=}10$ in Eq.~\eqref{eq:shift} unless otherwise noted.
Fig.~\ref{fig:teaser} instead starts from standard fine-tuning with $s{=}5$, the default of Wan2.2, so that it also shows the gain of the decision-focused schedule.
At inference, causal models decode new frames on the fly and stop at the first frame whose border pixels are at least 70\% red, with a maximum length of 20\,s.
Fixed-length baselines generate their training length, which matches the longest training video in each task.
Our models are sampled with 5 denoising steps, and baselines use their own samplers.
Full hyperparameters are listed in the supplement.

\subsection{Generation Paradigm}
\label{sec:exp_paradigm}

We first study how the generation process should be structured for multi-step reasoning.

\noindent\textbf{Baselines.}
We compare causal generation with adaptive termination against two groups of alternatives.
(1) \emph{Zero-shot video generators}: the closed-source Seedance 2.5~\cite{seedance25} and Veo 3.1~\cite{veo31}, the open-source Wan2.2-5B~\cite{wan}, and the reasoning-tuned VBVR-Wan2.2-A14B~\cite{vbvr} and VBVR-Pro-Wan2.2-5B-RLVR~\cite{vbvrpro}, which are trained on the VBVR task families.
(2) \emph{Fixed-length generation}: standard bidirectional SFT as in VBVR, which denoises all frames jointly without modifying the architecture; HDR~\cite{hdr}, which denoises a hierarchy of keyframes from coarse to fine; and S2PD~\cite{s2pd}, which switches from serial to parallel denoising.
HDR and S2PD are implemented on our backbones following their papers.
For \ours{}, we evaluate two segment lengths $K$ (Sec.~\ref{sec:causal_diffusion}): frame-wise generation with $K{=}1$ and chunk-wise generation with $K{=}8$.

\begin{table}[t]
\centering
\footnotesize
\setlength{\tabcolsep}{3.5pt}
\caption{\textbf{Per-task results on VBVR-MultiStep} (SSR, \%).}
\label{tab:paradigm_vbvr}
\begin{tabular*}{\linewidth}{@{\hspace{\tabcolsep}\extracolsep{\fill}}lcccccc}
\toprule
Method & Hanoi & Ordinal & Sliding & Snake & Sokoban & TSP \\
\midrule
\catrow{7}{Fixed-length generation}
Bidirectional & 8.0 & 2.0 & 9.0 & 5.0 & 39.0 & 71.0 \\
HDR~\cite{hdr} & 13.0 & 27.0 & 12.0 & 1.0 & 19.0 & 72.0 \\
S2PD~\cite{s2pd} & 21.0 & 94.0 & 52.0 & 70.0 & 64.0 & 86.0 \\
\catrow{7}{Causal generation with adaptive termination}
Ours ($K{=}8$) & 33.0 & 94.0 & 47.0 & 37.0 & 51.0 & 85.0 \\
Ours ($K{=}1$) & \textbf{37.0} & \textbf{97.0} & \textbf{61.0} & \textbf{86.0} & \textbf{75.0} & \textbf{93.0} \\
\bottomrule
\end{tabular*}
\end{table}

\begin{table}[t]
\centering
\footnotesize
\setlength{\tabcolsep}{4pt}
\caption{\textbf{Per-task results on RoboTwin-Plan} (SSR, \%).}
\label{tab:paradigm_robotwin}
\begin{tabular*}{\linewidth}{@{\hspace{\tabcolsep}\extracolsep{\fill}}lcccccc}
\toprule
& \multicolumn{2}{c}{Blocks World} & \multicolumn{2}{c}{Tower of Hanoi} & \multicolumn{2}{c}{Color Sorting} \\
\cmidrule(lr){2-3}\cmidrule(lr){4-5}\cmidrule(lr){6-7}
Method & Row & Triangle & Row & Triangle & Buffer & No buffer \\
\midrule
\catrow{7}{Fixed-length generation}
Bidirectional & 66.0 & 30.0 & 98.0 & 66.0 & 55.0 & 5.0 \\
HDR~\cite{hdr} & \textbf{86.0} & 56.7 & 90.0 & 72.0 & 75.0 & 35.0 \\
S2PD~\cite{s2pd} & 79.0 & 46.3 & 92.0 & 84.0 & 80.0 & 54.0 \\
\catrow{7}{Causal generation with adaptive termination}
Ours ($K{=}8$) & 85.0 & \textbf{64.0} & 98.0 & 74.0 & 90.0 & 64.0 \\
Ours ($K{=}1$) & 75.0 & 50.7 & \textbf{100} & \textbf{88.0} & \textbf{96.0} & \textbf{74.0} \\
\bottomrule
\end{tabular*}
\end{table}

\begin{table}[t]
\centering
\footnotesize
\caption{\textbf{Training schedule} (SSR, \%). A larger shift $s$ in Eq.~\eqref{eq:shift} moves more training to high noise levels; $s{=}1$ applies no shift, and $s{=}5$ is the default of Wan2.2.}
\label{tab:schedule}
\begin{tabular*}{\linewidth}{@{\hspace{\tabcolsep}\extracolsep{\fill}}lccc}
\toprule
Benchmark & $s{=}1$ & $s{=}5$ & $s{=}10$ \\
\midrule
\catrow{4}{Wan2.2-5B fine-tuned}
FlowFree & 46.8 & 73.5 & \textbf{77.9} \\
Sokoban & 68.4 & 79.6 & \textbf{83.0} \\
VBVR-MultiStep & 44.0 & 70.3 & \textbf{74.8} \\
RoboTwin-Plan & 2.6 & 33.3 & \textbf{73.9} \\
\catrow{4}{DiT-B trained from scratch}
FlowFree & 0.0 & 51.5 & \textbf{58.6} \\
Sokoban & 2.0 & 70.8 & \textbf{73.8} \\
\bottomrule
\end{tabular*}
\end{table}

\begin{table*}[t]
\centering
\small
\caption{\textbf{Comparison of post-training methods} (SSR, \%). The SFT model is our best result in Tab.~\ref{tab:paradigm}, and all methods with + start from it under the same training budget. S2PD is the strongest baseline in Tab.~\ref{tab:paradigm}.}
\label{tab:posttrain}
\setlength{\tabcolsep}{6.97pt}%
\begin{tabular}{lcccccccccc}
\toprule
& \multicolumn{6}{c}{Wan2.2-5B fine-tuned} & \multicolumn{4}{c}{DiT-B trained from scratch} \\
\cmidrule(lr){2-7}\cmidrule(lr){8-11}
& \multicolumn{2}{c}{FlowFree} & \multicolumn{2}{c}{Sokoban} & VBVR-MultiStep & RoboTwin-Plan & \multicolumn{2}{c}{FlowFree} & \multicolumn{2}{c}{Sokoban} \\
\cmidrule(lr){2-3}\cmidrule(lr){4-5}\cmidrule(lr){8-9}\cmidrule(lr){10-11}
Method & Test & Hard & Test & Hard & Test & Test & Test & Hard & Test & Hard \\
\midrule
S2PD~\cite{s2pd} & 68.1 & 41.2 & 75.7 & 38.2 & 64.5 & 62.9 & 41.2 & 25.6 & 56.4 & 11.6 \\  %
\rowcolor{oursblue!8} SFT model (ours) & 77.9 & 44.8 & 83.0 & 50.0 & 74.8 & 73.9 & 58.6 & 24.4 & 73.8 & 32.8 \\
+ Continued SFT & 75.5 & 48.6 & 85.5 & 59.6 & 74.2 & 74.3 & 55.8 & \textbf{34.4} & 74.1 & 42.4 \\
\catrow{11}{Sample-level}
+ Flow-GRPO~\cite{flowgrpo} & 72.9 & 47.0 & 80.6 & 45.4 & 73.2 & 73.4 & 57.7 & 31.8 & 73.0 & 37.2 \\
+ DiffusionNFT~\cite{diffusionnft} & 79.5 & 43.0 & 80.5 & 45.6 & 72.7 & 72.0 & 60.2 & 26.0 & 71.6 & 28.0 \\
+ RFT & 83.2 & 50.2 & 90.1 & 61.6 & 79.8 & 77.9 & 60.7 & 30.8 & 77.9 & 43.2 \\
\catrow{11}{Decision-level}
+ Flow-GRPO & 73.0 & 45.8 & 81.0 & 47.0 & 74.3 & 75.0 & 57.0 & 31.0 & 74.1 & 39.0 \\
+ DiffusionNFT & 73.8 & 43.2 & 78.6 & 40.0 & 67.2 & 63.3 & 61.4 & 25.0 & 72.8 & 28.6 \\
+ RFT & 80.1 & \textbf{52.0} & 89.8 & 73.8 & 78.8 & 70.7 & 59.0 & 28.0 & 79.1 & 46.3 \\
\rowcolor{oursblue!8} + CFT (ours) &\textbf{84.4} & \textbf{52.0} & \textbf{90.8} & \textbf{78.2} & \textbf{82.2} & \textbf{78.0} & \textbf{65.6} & 27.2 & \textbf{79.6} & \textbf{48.3} \\
\bottomrule
\end{tabular}
\vspace{0.5mm}
\end{table*}

\begin{figure}[t]
\centering
\includegraphics[width=\linewidth]{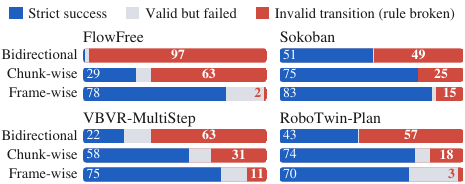}  %
\caption{\textbf{Failure analysis of generation paradigms} on Wan2.2-5B, with each test video categorized by its first error (\%).}
\label{fig:rule_breaks}
\end{figure}

\noindent\textbf{Results.}
Causal generation outperforms all alternatives on every benchmark and both backbones (Tab.~\ref{tab:paradigm}).
On FlowFree, for example, frame-wise causal generation reaches 77.9\%, while bidirectional SFT and the strongest structured baseline, S2PD, reach 1.1\% and 68.1\%.
Video models tuned on large-scale reasoning data also perform poorly, with VBVR-Wan2.2 and VBVR-Pro-RLVR solving only 11.3\% and 7.7\% of VBVR-MultiStep.
As Figs.~\ref{fig:qualitative} and~\ref{fig:rule_breaks} show, bidirectional generation fails to learn even the basic rules, and the vast majority of its failures are invalid state transitions.
In contrast, frame-wise causal generation learns valid state transitions and rarely breaks the rules.
Frame-wise generation works best on the puzzles, while chunk-wise generation is best on RoboTwin-Plan, mainly on Blocks World, likely because a single pick-and-place transition lasts about as long as one chunk.
Maze is omitted from Tab.~\ref{tab:paradigm}, as it does not require reasoning over state transitions and every paradigm exceeds 98\% after training.

\subsection{Training Schedule}
\label{sec:exp_schedule}

We next study how training should be distributed over noise levels.
We vary the shift $s$ with the same data and training budget, where $s{=}1$ recovers the original timestep distribution.
For DiT-B, whose timesteps follow a logit-normal distribution, the same shift moves its mean to $\log s$.
Each benchmark uses its best generation paradigm from Sec.~\ref{sec:exp_paradigm}: frame-wise generation on VideoRLVR and VBVR-MultiStep, and chunk-wise generation on RoboTwin-Plan.

\noindent\textbf{Results.}
Shifting training toward high noise levels improves SSR on every benchmark and both backbones (Tab.~\ref{tab:schedule}).
Without the shift, DiT-B trained from scratch solves almost no puzzle, whereas $s{=}10$ reaches 58.6\% and 73.8\%.
RoboTwin-Plan also requires a stronger shift than the default $s{=}5$ of Wan2.2, which reaches only 33.3\% against 73.9\% at $s{=}10$, likely because robot videos contain much more visual detail than the puzzles.
Longer training does not close this gap, as the model trained without the shift for 10 epochs (46.8\% on FlowFree and 68.4\% on Sokoban) remains well below $s{=}10$ trained for only 5 epochs (69.4\% and 79.6\%).

\subsection{RL Post-Training}
\label{sec:exp_cft}

\begin{figure*}[t]  %
\centering
\includegraphics[width=\linewidth]{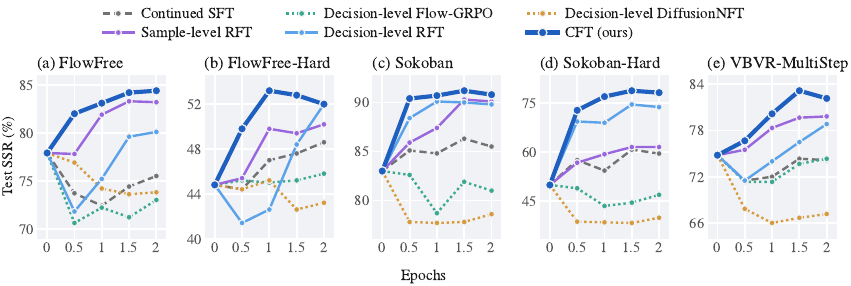}  %
\caption{\textbf{Test SSR during post-training} on Wan2.2-5B (\%), evaluated on the held-out test sets every half epoch. All methods start from the same SFT model at epoch 0.}
\label{fig:posttrain_curves}
\end{figure*}

Finally, we study how the SFT model can be further improved by learning from its own rollouts.

\noindent\textbf{Data.}
As the SFT models already solve 80--90\% of the VideoRLVR training puzzles, we add 5,000 harder ones to FlowFree and Sokoban and evaluate on two held-out hard sets of 500 puzzles, FlowFree-Hard and Sokoban-Hard.
On VBVR-MultiStep and RoboTwin-Plan, failures are collected only from the three and four sub-tasks with the lowest success, since the others are already largely solved.
The expert videos mixed into training still cover all sub-tasks.

\looseness=-1
\noindent\textbf{Baselines.}
All methods start from the best SFT models of Sec.~\ref{sec:exp_paradigm} and~\ref{sec:exp_schedule}, are trained with the same budget, and are evaluated at their final checkpoint.
Besides continued SFT on expert videos, we compare two families of methods that learn from the model's own rollouts, with strict success as the reward.
(1) \emph{Sample-level} methods roll out whole videos for each puzzle: Flow-GRPO~\cite{flowgrpo}, DiffusionNFT~\cite{diffusionnft}, and rejection-sampling fine-tuning (RFT)~\cite{raft,rft}, which fine-tunes on successful rollouts.
Sample-level Flow-GRPO corresponds to the RL used by prior video reasoning models such as VideoRLVR~\cite{videorlvr} and VBVR-Pro~\cite{vbvrpro}.
(2) \emph{Decision-level} methods instead branch from the decision where a rollout fails and apply the same three objectives to the resulting continuations.
In particular, decision-level RFT trains on the same successful continuations as CFT but with fresh noise and without the extra weight at $t{=}1$, which makes it an ablation of CFT without noise mapping.
Implementation details of all methods are provided in the supplement.

\noindent\textbf{Results.}
CFT further improves the SFT model by 4.1--7.8 points on the original test sets (Tab.~\ref{tab:posttrain}).
Continued SFT on expert videos brings no consistent gain and even lowers SSR on FlowFree and VBVR-MultiStep, so further progress requires learning from the model's own rollouts.
Flow-GRPO and DiffusionNFT bring little or no gain in these tasks.
Even when branching locates the failed decision, a scalar reward on a high-dimensional segment yields a noisy update.
Branching at decisions alone is not sufficient, as decision-level RFT underperforms sample-level RFT on most test sets and helps mainly on hard puzzles such as Sokoban-Hard.
Noise mapping turns these branches into targeted corrections. CFT improves over decision-level RFT on every original test set, by up to 7.3 points.
Sample-level RFT, the strongest baseline, comes close on the original test sets but lags far behind on Sokoban-Hard and learns much more slowly. As the curves in Fig.~\ref{fig:posttrain_curves} show, CFT already improves on every benchmark within half an epoch, \eg, by more than 20 points on Sokoban-Hard, while on FlowFree and VBVR-MultiStep the other methods are still near or below the SFT model at that point.

\begin{figure}[t]
\centering
\includegraphics[width=\linewidth]{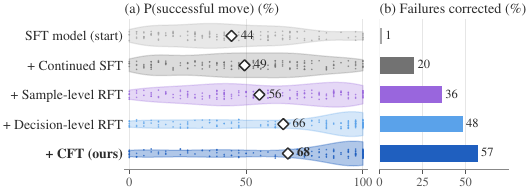}  %
\caption{\textbf{CFT shifts the decision distribution toward successful decisions.} We probe the 149 points in the Sokoban test sets where a failed rollout and a successful continuation diverge. (a) Probability of the successful decision over 32 fresh noises per point. (b) Percentage of divergence points where the noise of the failed rollout now yields the successful decision.}
\label{fig:mechanism}
\end{figure}

\noindent\textbf{Comparison with solver corrections.} On rule-based games, an exact solver can correct every failed rollout, including those for which no resampled continuation succeeds. Self-generated corrections require a successful resample and cover only about a third as many FlowFree puzzles. However, CFT reaches 84.4\% on FlowFree, while training on the solver corrections reaches only 48.2\%, as they lie outside the model's own distribution (see the supplement).

\noindent\textbf{Analysis.}
CFT is designed to redirect the noise of a failed rollout toward a successful decision and, through many such corrections, to shift the decision distribution toward successful decisions.
As shown in Fig.~\ref{fig:mechanism}, we probe both effects at 149 points in the Sokoban test sets where a failed rollout and a successful continuation diverge, by sampling the next decision from their shared history.
First, the noise of the failed rollout now yields the successful decision at 57.0\% of the points after CFT, compared with 48.3\% after decision-level RFT, which learns from the same continuations without noise mapping.
Second, the correction extends beyond that noise, as CFT chooses the successful decision with an average probability of 67.9\% over 32 fresh noises, compared with 65.9\% for decision-level RFT and 43.8\% for the SFT model.
At the task level, CFT also improves pass@$k$ at $k = 1, 2, 4$ on both games (see the supplement).

\begin{figure*}[t]  %
\centering
\includegraphics[width=\linewidth]{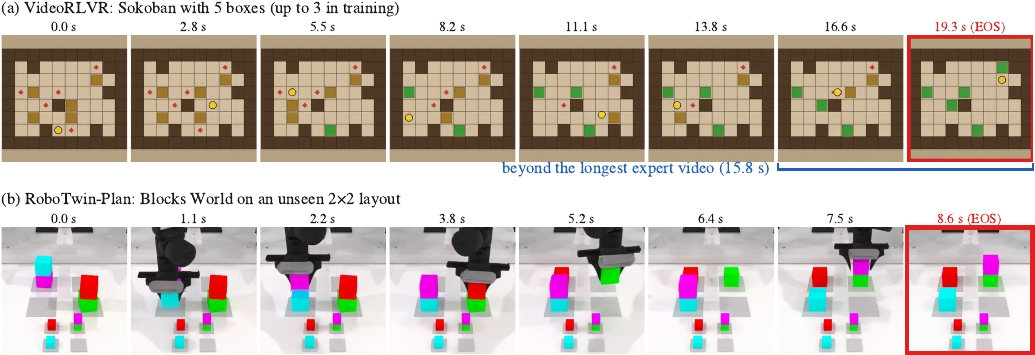}  %
\caption{\textbf{Out-of-distribution results.} Our models generalize to a larger puzzle (a) and a new layout (b).}
\label{fig:ood}
\end{figure*}

\noindent\textbf{Multi-round CFT.}
We can repeat CFT on the failures of the first-round CFT model.
As Tab.~\ref{tab:cft_analysis} shows, the second round improves both games, especially on the hard sets, whereas training the first-round model for the same number of updates on its original data brings little gain.
The improvement therefore comes from the new failures rather than from longer training.

\begin{table}[t]
\centering
\footnotesize
\setlength{\tabcolsep}{4pt}
\caption{\textbf{Multi-round CFT} (SSR, \%) on Wan2.2-5B.}
\label{tab:cft_analysis}
\begin{tabular*}{\linewidth}{@{\hspace{\tabcolsep}\extracolsep{\fill}}lcccc}
\toprule
& \multicolumn{2}{c}{FlowFree} & \multicolumn{2}{c}{Sokoban} \\
\cmidrule(lr){2-3}\cmidrule(lr){4-5}
Setting & Test & Hard & Test & Hard \\
\midrule
CFT, round 1 & 84.4 & 52.0 & 90.8 & 78.2 \\
+ same updates, round-1 data & 84.5 & 53.0 & 90.7 & 76.4 \\
+ round 2 & \textbf{86.2} & \textbf{57.6} & \textbf{92.4} & \textbf{80.2} \\
\bottomrule
\end{tabular*}
\end{table}

\subsection{Out-of-Distribution Generalization}
\label{sec:exp_ood}

\begin{table}[t]
\centering
\footnotesize
\setlength{\tabcolsep}{2.5pt}
\caption{\textbf{Out-of-distribution generalization} (SSR, \%) on Wan2.2-5B. CFT starts from the better causal model of each benchmark.}
\label{tab:ood}
\begin{tabular*}{\linewidth}{@{\hspace{\tabcolsep}\extracolsep{\fill}}lccccc}
\toprule
& \multicolumn{2}{c}{Sokoban} & FlowFree & \multicolumn{2}{c}{RoboTwin-Plan} \\
\cmidrule(lr){2-3}\cmidrule(lr){4-4}\cmidrule(lr){5-6}
Method & 4 boxes & 5 boxes & 9$\times$9 & Blocks 2$\times$2 & Hanoi 2$\times$2 \\  %
\midrule
Bidirectional & 0.0 & 2.0 & 0.0 & 14.0 & 16.0 \\
Ours ($K{=}8$) & 23.0 & 10.0 & 3.0 & \textbf{36.0} & \textbf{18.0} \\
Ours ($K{=}1$) & 51.0 & 37.0 & 32.0 & 4.0 & 2.0 \\
+ CFT (ours) & \textbf{74.0} & \textbf{54.0} & \textbf{38.0} & 32.0 & 14.0 \\  %
\bottomrule
\end{tabular*}
\end{table}

We also test whether the models generalize to larger puzzles and new layouts, as shown in Tab.~\ref{tab:ood} and Fig.~\ref{fig:ood}.

\noindent\textbf{Larger puzzles.}
We test 100 Sokoban puzzles each with 4 and 5 boxes (up to 3 in training) and 100 FlowFree puzzles on 9$\times$9 boards (up to 8$\times$8 in training).
Bidirectional generation solves almost none of them, while frame-wise causal generation solves 32--51\% of them, and CFT raises the success rate to 38--74\%.
On FlowFree, the solutions are longer than every training video, and the model still stops right after solving the board.

\noindent\textbf{New layouts.}
On RoboTwin-Plan, we apply trained rules to new layouts, with Blocks World on the 2$\times$2 layout and Tower of Hanoi on four pads instead of three.
Chunk-wise generation, the best paradigm on RoboTwin-Plan, solves 36.0\% of the Blocks World puzzles, compared with 14.0\% for bidirectional generation. CFT breaks the rules no more often than the chunk-wise model, but training on self-generated videos causes more visual defects, which slightly lower its success rate.
Together, these results suggest that our models learn transferable rules rather than only mimicking the training solutions.

%% file: sec/5_conclus.tex
\section{Conclusion}
\label{sec:conclus}

We presented \ours{}, a decision-centric framework that teaches video diffusion models to reason and plan step by step.
\ours{} generates a solution in causal order and decides when to stop, concentrates training on the high noise levels where decisions are made, and corrects its own failures with CFT.
Across abstract games, planning tasks, and robot manipulation, \ours{} raises the average strict success rate of standard fine-tuning from 23.3\% to 83.8\% and generalizes to larger puzzles and unseen layouts beyond the training range.
Despite its strong performance, \ours{} is currently trained on one benchmark at a time and relies on rule verifiers for post-training.
In the future, we plan to extend it to broader reasoning data, learned verifiers, and real-world robot planning.

%% file: sec/X_suppl.tex
\clearpage
\maketitlesupplementary
\setcounter{section}{0}
\setcounter{figure}{0}
\setcounter{table}{0}
\renewcommand{\thesection}{\Alph{section}}
\renewcommand{\thefigure}{S\arabic{figure}}
\renewcommand{\thetable}{S\arabic{table}}

This supplementary material provides further details and results that complement the main paper.
Sec.~\ref{supp:benchmarks} describes the datasets and benchmarks, and Sec.~\ref{supp:evaluation} explains how we evaluate the generated videos.
Sec.~\ref{supp:impl} presents the implementation details of our models, the baselines, and post-training.
Sec.~\ref{supp:results} and Sec.~\ref{supp:qualitative} report additional quantitative and qualitative results, respectively, and Sec.~\ref{supp:limitations} discusses limitations and future work.

\input{sec/X_suppl_data}

\input{sec/X_suppl_impl}

\input{sec/X_suppl_results}

%% file: sec/X_suppl_data.tex
\section{Datasets and Benchmarks}
\label{supp:benchmarks}

\begin{figure*}[t]
\centering
\includegraphics[width=\linewidth]{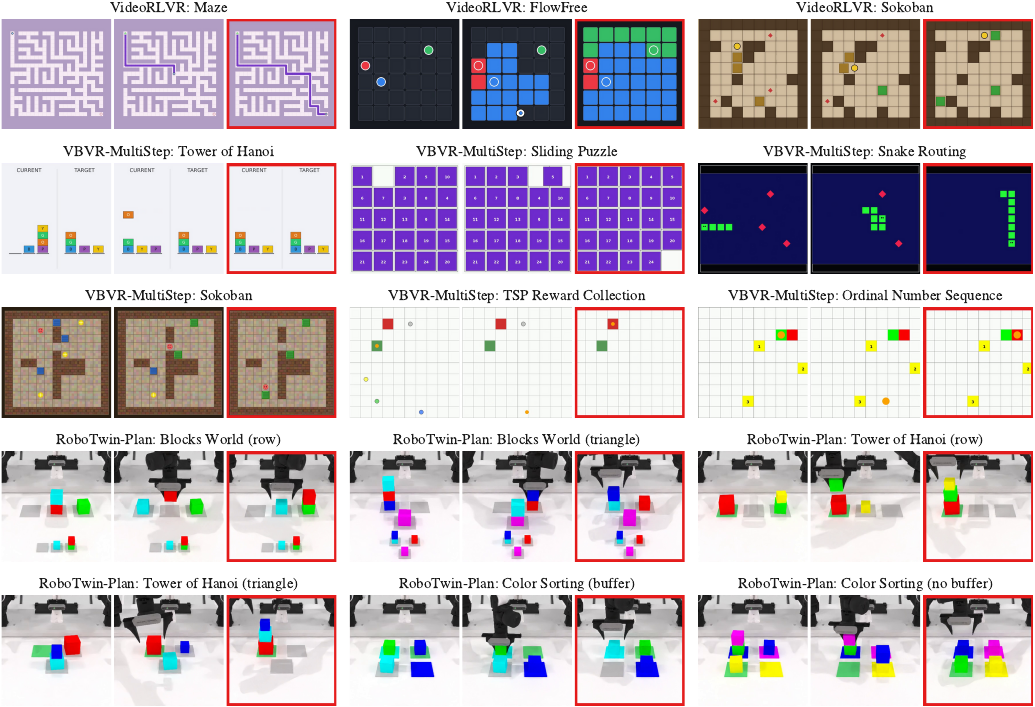}  %
\caption{\textbf{Test examples of the three benchmarks.} For each task, we show the first frame, the middle frame, and the last frame of the reference video. Every video ends with EOS frames that repeat the final state with a red border.}
\label{fig:supp_benchmarks}
\end{figure*}

Our experiments use three benchmarks that range from abstract games to robot manipulation: the rule-based games of VideoRLVR, the planning tasks of VBVR-MultiStep, and RoboTwin-Plan, which we build in the RoboTwin 2.0 simulator.
Fig.~\ref{fig:supp_benchmarks} shows a test example of every task, Tab.~\ref{tab:supp_benchmarks} summarizes the data, and Tab.~\ref{tab:supp_prompts} lists the prompts.
For Wan2.2-5B, all videos have $384{\times}384$ pixels and 16\,fps with their native duration, and each ends with eight frames that hold the final state, the last four of which carry a 10-pixel red border as the visual EOS.
DiT-B uses $256{\times}256$ pixel videos of the games with one frame per move (Sec.~\ref{supp:impl_dit}).

\begin{table*}[t]
\centering
\small
\setlength{\tabcolsep}{5pt}
\caption{\textbf{Benchmark statistics.} The setting describes the test puzzles, and the frame counts are the median and maximum over the training videos. VideoRLVR holds out 500 of its 10K training puzzles per game for validation.}
\label{tab:supp_benchmarks}
\begin{tabular*}{\linewidth}{@{\hspace{\tabcolsep}\extracolsep{\fill}}llrrlc}
\toprule
Benchmark & Task & Train & Test & Setting & Frames \\
\midrule
\multirow{3}{*}{VideoRLVR} & Maze & 9,500 & 1,000 & 5--15 passages per side & 77 / 105 \\
& FlowFree & 9,500 & 1,000 & $5{\times}5$--$8{\times}8$ boards, 3--5 colors & 101 / 173 \\
& Sokoban & 9,500 & 1,000 & 1--3 boxes, 2--51 moves & 69 / 253 \\
\midrule
\multirow{6}{*}{VBVR-MultiStep} & Tower of Hanoi & 5,000 & 100 & 3--5 blocks, 1--11 moves & 97 / 193 \\
& Ordinal Number Sequence & 5,000 & 100 & 3 numbers & 89 / 177 \\
& Sliding Puzzle & 5,000 & 100 & $3{\times}3$--$5{\times}5$ boards, 7--15 moves & 69 / 105 \\
& Snake Routing & 5,000 & 100 & $10{\times}10$--$14{\times}14$ grids, 3--5 foods & 57 / 145 \\
& Sokoban & 5,000 & 100 & 1--3 boxes, 4--40 moves & 41 / 129 \\
& TSP Reward Collection & 5,000 & 100 & $10{\times}10$ board, 3--4 rewards & 81 / 121 \\
\midrule
\multirow{6}{*}{RoboTwin-Plan} & Blocks World (row) & 2,000 & 100 & 3--4 blocks, 2--4 moves & 85 / 121 \\
& Blocks World (triangle) & 5,000 & 300 & 3--4 blocks, 2--6 moves & 121 / 185 \\
& Tower of Hanoi (row) & 500 & 50 & 3 blocks, 2--4 moves & 89 / 121 \\
& Tower of Hanoi (triangle) & 500 & 50 & 3 blocks, 2--5 moves & 113 / 161 \\
& Color Sorting (buffer) & 1,000 & 100 & 3 blocks, 2--4 moves & 85 / 125 \\
& Color Sorting (no buffer) & 1,000 & 100 & 4 blocks, 2--6 moves & 137 / 177 \\
\bottomrule
\end{tabular*}
\end{table*}

\noindent\textbf{VideoRLVR.}
We parse every puzzle of VideoRLVR~\cite{videorlvr} into a symbolic state and render it again with one renderer for all games.
Maze videos draw the path at 1.5 frames per move, and Sokoban videos show each move or push over four frames.
FlowFree videos fill one cell at a time along a single continuous walk, which we recover from the drawing order of the original videos.
The walk only switches to a new color at one of its endpoints, after every endpoint of the current color has been visited, and that endpoint must be adjacent to the previous cell.
Since the original test set contains no $8{\times}8$ boards and some FlowFree puzzles fail the conversion, we replace them with new puzzles generated following the construction of VideoRLVR, which builds a Hamiltonian path over the board with Warnsdorff's rule and cuts it into contiguous color segments.
The FlowFree test set thus contains 123 generated puzzles, including all 100 $8{\times}8$ boards.

\noindent\textbf{Hard sets and post-training puzzles.}
FlowFree-Hard contains 500 boards of $5{\times}5$ to $8{\times}8$ cells with 4--6 colors, generated in the same way.
Sokoban-Hard contains 500 puzzles with two or three boxes in new rooms of $7{\times}7$ to $9{\times}9$ cells including walls, each with a move-optimal solution of 16--31 moves found by A$^*$ search.
For post-training, we generate 5,000 additional training puzzles for each game with the same generators.
All layouts of the hard sets are disjoint from VideoRLVR and from these training puzzles.

\noindent\textbf{VBVR-MultiStep.}
We take the videos and prompts of the six planning tasks of VBVR-MultiStep\footnote{\url{https://huggingface.co/datasets/Video-Reason/VBVR-MultiStep}}~\cite{vbvr}, remove idle frames, and resize the videos to $384{\times}384$.
Once the colors are ignored, the original test split shares many layouts with the training split, \eg, 45\% of the Tower of Hanoi test puzzles, so we build a clean test set of 100 puzzles per task whose color-free layouts never occur in training.
Except for Sokoban, the prompts ask for the shortest solution, \eg, ``the minimum total number of moves'' for Tower of Hanoi, whose blocks have equal sizes and are rearranged from a current to a target configuration rather than stacked as in the classical puzzle.

\noindent\textbf{RoboTwin-Plan.}
We build RoboTwin-Plan in the RoboTwin 2.0 simulator\footnote{\url{https://github.com/RoboTwin-Platform/RoboTwin}}~\cite{robotwin} with the dual-arm Aloha-AgileX robot, a clean table, and a fixed third-person camera.
Blocks World uses three or four cubes on three places in a row or a triangle, and the goal arrangement is shown by a small model at the front of the table.
Tower of Hanoi uses three blocks of different sizes on three pads, which must all be stacked on the green pad without placing a larger block on a smaller one.
Color Sorting uses a $2{\times}2$ layout of colored pads, where each block must end alone on the pad of its color, either with a free gray buffer pad for three blocks or without one for four blocks.
For each episode, we compute the shortest symbolic plan with breadth-first search and execute it with the grasp and place primitives of RoboTwin, moving one top block at a time.
An episode is kept only if the simulator reaches the goal arrangement and the rendered video passes a replay of the symbolic plan.
Videos are rendered with ray tracing at one frame per 90 simulation steps, and one pick-and-place lasts about 25 frames.
The three splits contain no identical episode.

\noindent\textbf{Out-of-distribution sets.}
\label{supp:ood_sets}
The Sokoban sets place four or five boxes, targets, and the agent at random and keep puzzles with move-optimal solutions of at most 40 moves, which fit the fixed length of bidirectional generation.
The FlowFree set contains 100 $9{\times}9$ boards with 3--5 colors, whose videos last 213 frames, longer than every FlowFree training video.
On RoboTwin-Plan, Blocks World on the $2{\times}2$ layout uses the four places of Color Sorting, and Tower of Hanoi on four pads uses the same layout.
The longer plans in Blocks World and Tower of Hanoi require more moves than any training puzzle of the same task.
All reference videos of these sets pass the verifiers.

\section{Evaluation}
\label{supp:evaluation}

As noted in Sec.~\ref{sec:exp_setup}, the original evaluations of these benchmarks do not follow the task rules exactly.
VideoRLVR judges Maze and FlowFree on the final frame only and accepts a Sokoban video with up to half of its transitions invalid.
Its precision, recall, and F1 also compare a video with a single reference solution and thus penalize other valid solutions.
VBVR gives partial credit based on pixel heuristics, and VBVR-MultiStep has no automatic evaluation.
Therefore, we build a unified, automatic evaluation pipeline for the Strict Success Rate (SSR) on all benchmarks.
It first detects when each generated video terminates, and then replays its states with a verifier that checks every transition under the task rules.
Following prior work in which LLMs write reward functions as code~\cite{eureka,text2reward}, we build the verifiers agentically.
A coding agent implements the verifier of each task from a few general rules shared by all its instances and refines it until it accepts every reference video.
We then validate the final verifiers against human judgments on generated videos.

\noindent\textbf{Termination.}
\label{supp:verifiers}
The pipeline begins by detecting the EOS frame of each generated video.
A frame counts as EOS if at least 70\% of its border pixels are red, and all frames from the first EOS frame on are ignored.
Causal models stop at this frame or at the maximum length limit of 20\,s, and fixed-length models generate their full length.

\noindent\textbf{Verifiers.}
Every verifier reads the state of each frame, replays the trajectory under the task rules, and reports the first violation.
\begin{itemize}
\item \emph{FlowFree} reads the color of every cell from its interior pixels.
A newly painted cell must extend the current color from its tip or start a new color at one of its endpoints next to the previous cell, after all endpoints of the current color have been visited.
Paint may never disappear or change color.
The video succeeds when the board is fully painted without any earlier violation.
\item \emph{Sokoban} parses the agent, the boxes, and the targets in every frame.
It rejects hallucinated or vanished objects and requires every change between consecutive settled states to be a legal move or push.
Two moves within one transition are accepted only if their intermediate state appears in a frame.
The video must end with all boxes on targets.
\item \emph{Maze} requires the drawn path to stay off the walls and remain connected to the start, and the final path to connect the start and the goal.
\item \emph{VBVR-MultiStep} reads a symbolic state per frame with the geometry of each puzzle, and its task-specific rules reject illegal moves and hallucinated, vanished, or teleporting objects.
The shortest-solution requirement compares the number of moves, or the path length for TSP Reward Collection and Ordinal Number Sequence, with the reference solution.
\item \emph{RoboTwin-Plan} uses a learned state reader, a frozen DINOv3~\cite{dinov3} encoder with a transformer decoder trained on the simulator states of the training videos, which predicts the position of every block and whether each gripper holds it.
A Viterbi decoder turns these predictions into a sequence of arrangements, and the rules reject blocks that move without being grasped, duplicated or vanished blocks, and every change between rest states that is not one legal move of a top block, including the size rule of Tower of Hanoi.
A separate appearance classifier on DINOv3 patch features flags local visual defects of the generated frames.
\end{itemize}
We validate the verifiers on reference and generated videos.
All reference videos of the game and VBVR-MultiStep test sets pass, and the RoboTwin-Plan rules accept all 540 expert videos we tested.
The RoboTwin-Plan state reader identifies the held block with 99.9\% accuracy and localizes blocks with a median error of 0.6\,mm on held-out expert videos, and the full verifier agrees with human labels on over 95\% of 140 generated videos.

\noindent\textbf{Strict and relaxed success.}
A video can reach the goal state while breaking the rules on the way, and Tab.~\ref{tab:supp_strict} measures how often this happens.
On Sokoban, bidirectional generation reaches the goal state in 98.2\% of the videos but follows every rule in only 50.8\%, whereas frame-wise causal generation loses only 0.5 points under the strict metric.
The extra goal-reaching videos of bidirectional generation are not solutions, as they break the rules on the way, \eg, boxes vanish or reappear, the agent duplicates, or objects teleport.
On VBVR-MultiStep, the shortest-solution requirement mainly affects Tower of Hanoi, and the ranking of the paradigms is the same with and without it.

\noindent\textbf{Zero-shot generators.}
Since the closed-source models are costly to query and almost never succeed under the strict rules, we evaluate them on stratified test subsets.
The subsets contain 20 FlowFree puzzles (five per board size), 20 Sokoban puzzles (seven, seven, and six with one, two, and three boxes), 30 VBVR-MultiStep puzzles (five per task), and 30 RoboTwin-Plan episodes (five per sub-task).
Each model receives the first frame and the task prompt used in training.
Seedance 2.5 generates the length of the reference solution rounded up to whole seconds, between 4 and 20\,s, and Veo 3.1 generates its fixed 8\,s.
The open models, Wan2.2-5B, VBVR-Wan2.2-A14B\footnote{\url{https://huggingface.co/Video-Reason/VBVR-Wan2.2}}, and VBVR-Pro-Wan2.2-5B-RLVR\footnote{\url{https://huggingface.co/Video-Reason/VBVR-Pro-Wan2.2-TI2V-5B-RLVR}}, use their default sampling settings and are evaluated on 100 FlowFree puzzles (25 per board size), 100 Sokoban puzzles, 120 RoboTwin-Plan episodes (20 per sub-task), and the VBVR-MultiStep test set.
All outputs are resized to $384{\times}384$, and appended with four EOS frames before the verifiers are applied.
Since these models are not trained to stop or to find the shortest solution, we do not require the optimal number of moves.
As our verifiers are mainly built for the rendering and pacing of our datasets, which these models might not follow, we also review the videos by hand, relax the rules, and accept those whose errors are minor.

\begin{table}[t]
\centering
\footnotesize
\setlength{\tabcolsep}{3pt}
\caption{\textbf{Strict and relaxed success} (\%) of the generation paradigms on Wan2.2-5B, scored on the same videos. ``End state'' only checks the final frame, and ``strict'' additionally checks every transition. On VBVR-MultiStep, ``shortest'' additionally requires the optimal number of moves where the prompt asks for it.}
\label{tab:supp_strict}
\begin{tabular*}{\linewidth}{@{\hspace{\tabcolsep}\extracolsep{\fill}}lcccc}
\toprule
& \multicolumn{2}{c}{Sokoban} & \multicolumn{2}{c}{VBVR-MultiStep} \\
\cmidrule(lr){2-3}\cmidrule(lr){4-5}
Method & End state & Strict & Strict & + Shortest \\
\midrule
Bidirectional & 98.2 & 50.8 & 24.3 & 22.3 \\
Ours ($K{=}8$) & 84.5 & 75.0 & 65.2 & 57.8 \\
Ours ($K{=}1$) & 83.5 & 83.0 & 85.7 & 74.8 \\
\bottomrule
\end{tabular*}
\end{table}

%% file: sec/X_suppl_impl.tex
\section{Implementation Details}
\label{supp:impl}

This section gives the training and inference details of our models on Wan2.2-5B and DiT-B, of the baselines, and of post-training.

\subsection{Causal Generation on Wan2.2-5B}
\label{supp:impl_causal}

We fine-tune all 5.0B parameters of the diffusion transformer of Wan2.2-TI2V-5B\footnote{\url{https://huggingface.co/Wan-AI/Wan2.2-TI2V-5B}}~\cite{wan} and keep the text encoder and the VAE frozen.
The VAE encodes every four frames into one latent frame of $24{\times}24{\times}48$, which yields 144 tokens, and the first latent frame encodes the initial observation $\mathbf{o}_0$.
A segment of $K$ latent frames thus spans 4 video frames in frame-wise generation ($K{=}1$) and 32 in chunk-wise generation ($K{=}8$).

\noindent\textbf{Training.}
We train with teacher forcing on a clean and a noisy copy of each video.
Each noisy segment draws its own noise level from $p_{\mathrm{DF}}(t)$ and attends to its own noisy tokens and to the clean tokens of all earlier segments.
The loss is the flow-matching objective with the timestep weighting of Wan2.2, and the first latent frame is excluded from it.
The shift $s$ of Eq.~\eqref{eq:shift} only changes the training distribution, so all models share one sampling schedule.

\noindent\textbf{Inference.}
The model generates one segment at a time from a key-value cache of the clean history, with five UniPC steps, a sampling shift of 5, and no classifier-free guidance.
After each segment, we decode it and check for the EOS border, and generation stops at the first EOS frame or after 80 latent frames, \ie, 20\,s.

\begin{table*}[t]
\centering
\small
\setlength{\tabcolsep}{5pt}
\caption{\textbf{SFT settings.} All paradigms of a benchmark share these settings, and DiT-B is trained on FlowFree and Sokoban.}
\label{tab:supp_sft}
\begin{tabular*}{\linewidth}{@{\hspace{\tabcolsep}\extracolsep{\fill}}lcccc}
\toprule
& \multicolumn{3}{c}{Wan2.2-5B} & DiT-B \\
\cmidrule(lr){2-4}\cmidrule(lr){5-5}
& VideoRLVR & VBVR-MultiStep & RoboTwin-Plan & VideoRLVR \\
\midrule
Updates & 47,500 & 37,500 & 16,700 & 11,000 \\
Batch & 2 & 4 & 6 & 32 \\
Epochs & 10 & 5 & 10 & 37 \\
Peak learning rate & $1{\times}10^{-5}$ & $1{\times}10^{-5}$ & $1{\times}10^{-5}$ & $1{\times}10^{-4}$ \\
Warmup / schedule & 500 / cosine & 500 / cosine & 500 / cosine & 500 / constant \\
AdamW $\beta$ & (0.9, 0.999) & (0.9, 0.999) & (0.9, 0.999) & (0.9, 0.95) \\
Weight decay & 0.01 & 0.01 & 0.01 & 0 \\
EMA decay & -- & -- & -- & 0.999 \\
\bottomrule
\end{tabular*}
\vspace{-2mm}
\end{table*}

\noindent\textbf{Optimization.}
Tab.~\ref{tab:supp_sft} lists the optimizer and schedule settings, and all models train with gradient clipping at 1.0 and bf16 mixed precision.
On FlowFree and Sokoban, training has two stages of five epochs, and the second stage keeps the optimizer state and restarts the cosine schedule with a peak of $3{\times}10^{-6}$ and 1,500 warmup updates.

\subsection{DiT-B}
\label{supp:impl_dit}

Following S2PD~\cite{s2pd}, we also implement our designs on DiT-B~\cite{dit} to show that they do not depend on a specific backbone.
DiT-B is a 132M-parameter diffusion transformer with width 768, depth 12, and 12 heads, which operates on $256{\times}256$ frames with $32{\times}32$ patches, 3D rotary position embeddings, and adaLN-zero conditioning on the noise level, without text input.
Its videos show one frame per move and end with one EOS frame.
It predicts the clean frame, and the loss is weighted by $1/\max(t, 0.05)^2$ with $t$ drawn from a logit-normal distribution with mean $\log s$ and unit variance.
Its learning rate decays linearly to zero over the last 1,000 updates, and we evaluate the EMA weights.
Sampling uses five Euler steps per frame on a uniform noise grid with a key-value cache.

\subsection{Baselines}
\label{supp:impl_baselines}

\noindent\textbf{Bidirectional generation.}
The bidirectional baseline uses the original Wan2.2-5B or DiT-B architecture and denoises all frames of a clip jointly with a single noise level, with the first latent frame fixed to the initial observation.
Every training video is extended to a fixed length by holding its final EOS frame, and this length equals the longest training video of each task.

\noindent\textbf{HDR and S2PD.}
We implement both methods on our backbones following their papers and train them with the same data, shift, and number of training videos as the bidirectional baseline.
On FlowFree and Sokoban, their second training stage starts from the first-stage weights with a new optimizer state.
HDR~\cite{hdr} denoises a hierarchy of 1, 2, 4, 8, 16, and 32 keyframes and finally all latent frames, each level conditioned on the coarser ones, and we sample it with 5, 8, 13, 20, 32, 50, and 50 Euler steps from the coarsest to the finest level.
S2PD~\cite{s2pd} trains on a mixture of single-frame blocks and whole-video blocks with a noised context, and its sampler first generates frames one at a time down to noise level 0.6 and then denoises all frames in parallel, with 10 Euler steps in total.

\subsection{Post-Training}
\label{supp:impl_posttrain}

\looseness=-1
\noindent\textbf{Collecting corrections.}
We roll out the SFT model twice on each training puzzle, including the 5,000 harder puzzles of each game, locate the first error of every failed rollout with the verifier, and resample four continuations from a branch point shortly before the error, keeping the successful ones.
On FlowFree, the branch point lies at most four moves before the fatal move.
On Sokoban, it lies four latent frames before the first state that an exact solver proves unsolvable.
On VBVR-MultiStep, it is the last settled state before the first violation, redundant move, unsolvable state, or early stop.
On RoboTwin-Plan, it is the rest state before the first violation or detour.
For each successful continuation, the divergence step $d$ is the first segment at which it departs from the failed rollout.
Tab.~\ref{tab:supp_posttrain} lists the numbers of corrections.
On VBVR-MultiStep and RoboTwin-Plan, failures are collected on the three and four sub-tasks with the lowest success.

\begin{table*}[t]
\centering
\small
\setlength{\tabcolsep}{5pt}
\caption{\textbf{Post-training settings.} ``Corrections'' is the number of successful continuations used for post-training, and the number in parentheses counts the training puzzles, or RoboTwin-Plan episodes, from which they are collected.}
\label{tab:supp_posttrain}
\begin{tabular*}{\linewidth}{@{\hspace{\tabcolsep}\extracolsep{\fill}}lcccccc}
\toprule
& \multicolumn{4}{c}{Wan2.2-5B} & \multicolumn{2}{c}{DiT-B} \\
\cmidrule(lr){2-5}\cmidrule(lr){6-7}
& FlowFree & Sokoban & VBVR-MultiStep & RoboTwin-Plan & FlowFree & Sokoban \\
\midrule
Start model & $K{=}1$ & $K{=}1$ & $K{=}1$ & $K{=}8$ & $K{=}1$ & $K{=}1$ \\
Corrections & 8,483 (2,257) & 6,962 (2,747) & 10,937 (3,932) & 3,430 (1,258) & 9,633 (3,128) & 5,791 (1,802) \\
Updates & 7,600 & 6,200 & 9,800 & 3,200 & 1,000 & 1,000 \\
Batch & 3 & 3 & 3 & 3 & 32 & 32 \\
Peak learning rate & $1{\times}10^{-6}$ & $1{\times}10^{-6}$ & $1{\times}10^{-6}$ & $1{\times}10^{-6}$ & $5{\times}10^{-6}$ & $5{\times}10^{-6}$ \\
Warmup / schedule & 1,000 / cosine & 1,000 / cosine & 1,000 / cosine & 800 / cosine & 500 / constant & 500 / constant \\
Expert share & 25\% & 25\% & 26\% & 29\% & 25\% & 25\% \\
\bottomrule
\end{tabular*}
\vspace{1mm}
\end{table*}

\noindent\textbf{CFT.}
CFT starts from the full training state of the SFT model and trains for about two epochs over the corrections.
Each training row is a successful continuation together with its history, and half of its loss weight lies on the divergence segment $\mathbf{X}^+_d$, where the corrective loss of Eq.~\eqref{eq:cft_corr} reuses the noise $\boldsymbol{\epsilon}^-_d$ of the failed rollout and draws $t{=}1$ with probability 0.5.
The other half covers the remaining segments with the decision-focused objective.
Expert videos fill a quarter of each batch, which sets $\lambda$ in Eq.~\eqref{eq:cft}.
On DiT-B, we additionally keep an EMA of the weights with a decay of 0.99.

\noindent\textbf{Baselines.}
All post-training baselines start from the same SFT model and use the same number of updates and learning-rate schedule as CFT on each benchmark.
Decision-level RFT trains on exactly the corrections of CFT with fresh noise at the divergence segment and samples its timesteps only from $p_{\mathrm{DF}}(t)$, without the extra weight at $t{=}1$.
Sample-level RFT trains on successful whole rollouts of the SFT model together with expert videos.
Continued SFT trains on expert videos of the same puzzles.
The online methods alternate collection and training in rounds, eight rounds on FlowFree and VBVR-MultiStep, six on Sokoban, and four on RoboTwin-Plan.
Each round of the sample-level variants rolls out four videos for each puzzle in a new group of training puzzles, and each round of the decision-level variants branches four continuations at the same number of failures, located with the rule of CFT.
The reward is the strict success of the verifier, including the shortest-solution requirement on VBVR-MultiStep, and expert videos fill a quarter of each batch.
Flow-GRPO~\cite{flowgrpo} uses one stochastic sampling step with a noise level of 0.7, a clipping range of 0.01, and a KL coefficient of 0.04.
DiffusionNFT~\cite{diffusionnft} uses a KL coefficient of $10^{-4}$, and we scale its loss by 0.02 on Wan2.2-5B and 0.03 on DiT-B so that its gradient norm matches that of CFT.
Otherwise the success rate of its rollouts drops quickly within three rounds.

%% file: sec/X_suppl_results.tex
\section{Additional Quantitative Results}
\label{supp:results}

We complement Sec.~\ref{sec:exp} with results by puzzle size, a comparison of teacher forcing and diffusion forcing, per-task and pass@$k$ results of post-training, details of the mechanism analysis, ablations and extensions of CFT, further out-of-distribution results, and a comparison with frontier multimodal language models.

\subsection{Results by Puzzle Size}
\label{supp:difficulty}

\begin{figure}[t]
\centering
\includegraphics[width=\linewidth]{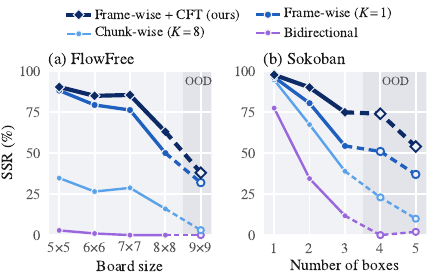}  %
\caption{\textbf{SSR by puzzle size} on Wan2.2-5B. FlowFree is grouped by board size and Sokoban by the number of boxes. The shaded sizes are not seen in training (Sec.~\ref{supp:ood_sets}).}
\label{fig:supp_difficulty}
\end{figure}

\begin{table*}[t]
\centering
\small
\setlength{\tabcolsep}{5pt}
\caption{\textbf{pass@$k$ after post-training} (SSR, \%) on Wan2.2-5B. A puzzle counts as solved if at least one of $k$ samples, drawn with $k$ fixed seeds, is strictly successful (on VBVR-MultiStep, strictly successful and shortest, as in Tab.~\ref{tab:posttrain}).}
\label{tab:supp_passk}
\begin{tabular*}{\linewidth}{@{\hspace{\tabcolsep}\extracolsep{\fill}}lcccccccccccc}
\toprule
& \multicolumn{3}{c}{FlowFree} & \multicolumn{3}{c}{FlowFree-Hard} & \multicolumn{3}{c}{Sokoban} & \multicolumn{3}{c}{Sokoban-Hard} \\
\cmidrule(lr){2-4}\cmidrule(lr){5-7}\cmidrule(lr){8-10}\cmidrule(lr){11-13}
Method & $k{=}1$ & $k{=}2$ & $k{=}4$ & $k{=}1$ & $k{=}2$ & $k{=}4$ & $k{=}1$ & $k{=}2$ & $k{=}4$ & $k{=}1$ & $k{=}2$ & $k{=}4$ \\
\midrule
SFT model (ours) & 77.9 & 90.8 & 96.1 & 44.8 & 59.4 & 72.2 & 83.0 & 88.7 & 92.0 & 50.0 & 64.2 & 73.8 \\
+ Continued SFT & 75.5 & 89.6 & 94.6 & 48.6 & 60.4 & 73.4 & 85.5 & 90.3 & 93.2 & 59.6 & 73.0 & 82.2 \\
+ Sample-level RFT & 83.2 & 92.9 & 96.6 & 50.2 & 63.0 & 76.6 & 90.1 & 93.1 & 94.8 & 61.6 & 75.2 & 83.4 \\
+ Decision-level RFT & 80.1 & 91.7 & 96.6 & \textbf{52.0} & 67.2 & \textbf{80.0} & 89.8 & 94.0 & 96.2 & 73.8 & 83.4 & 88.0 \\
+ CFT (ours) & \textbf{84.4} & \textbf{93.2} & \textbf{97.4} & \textbf{52.0} & \textbf{68.6} & 79.8 & \textbf{90.8} & \textbf{94.5} & \textbf{96.8} & \textbf{78.2} & \textbf{85.0} & \textbf{88.6} \\
\bottomrule
\end{tabular*}
\vspace{4pt}
\begin{tabular*}{\linewidth}{@{\hspace{\tabcolsep}\extracolsep{\fill}}lcccccc}
\toprule
& \multicolumn{3}{c}{VBVR-MultiStep} & \multicolumn{3}{c}{RoboTwin-Plan} \\
\cmidrule(lr){2-4}\cmidrule(lr){5-7}
Method & $k{=}1$ & $k{=}2$ & $k{=}4$ & $k{=}1$ & $k{=}2$ & $k{=}4$ \\
\midrule
SFT model (ours) & 74.8 & 82.7 & 87.8 & 73.9 & 84.3 & 89.7 \\
+ Continued SFT & 74.2 & 81.5 & 87.2 & 74.3 & 83.6 & \textbf{91.1} \\
+ Sample-level RFT & 79.8 & 86.0 & 90.5 & 77.9 & \textbf{86.3} & 90.9 \\
+ Decision-level RFT & 78.8 & 86.8 & 90.7 & 70.7 & 80.9 & 87.6 \\
+ CFT (ours) & \textbf{82.2} & \textbf{87.5} & \textbf{91.3} & \textbf{78.0} & 85.9 & 90.7 \\
\bottomrule
\end{tabular*}
\end{table*}

\begin{table}[t]
\centering
\footnotesize
\setlength{\tabcolsep}{2.0pt}
\caption{\textbf{Post-training on VBVR-MultiStep, per task} (SSR, \%, strict and shortest). The last column is the overall SSR of Tab.~\ref{tab:posttrain}.}
\label{tab:supp_vbvr_posttrain}
\begin{tabular*}{\linewidth}{@{\hspace{\tabcolsep}\extracolsep{\fill}}lccccccc}
\toprule
Method & Hanoi & Ordinal & Sliding & Snake & Sokoban & TSP & All \\
\midrule
SFT model (ours) & 37.0 & 97.0 & 61.0 & 86.0 & 75.0 & 93.0 & 74.8 \\
+ Continued SFT & 32.0 & 97.0 & 62.0 & 85.0 & 75.0 & 94.0 & 74.2 \\
\catrow{8}{Sample-level}
+ Flow-GRPO & 30.0 & 97.0 & 59.0 & 87.0 & 70.0 & \textbf{96.0} & 73.2 \\
+ DiffusionNFT & 32.0 & 96.0 & 57.0 & 85.0 & 72.0 & 94.0 & 72.7 \\
+ RFT & 54.0 & 97.0 & 69.0 & 87.0 & 79.0 & 93.0 & 79.8 \\
\catrow{8}{Decision-level}
+ Flow-GRPO & 34.0 & 97.0 & 60.0 & 89.0 & 72.0 & 94.0 & 74.3 \\
+ DiffusionNFT & 31.0 & 97.0 & 31.0 & 82.0 & 69.0 & 93.0 & 67.2 \\
+ RFT & 51.0 & 97.0 & \textbf{70.0} & 87.0 & 76.0 & 92.0 & 78.8 \\
+ CFT (ours) & \textbf{61.0} & \textbf{98.0} & 68.0 & \textbf{90.0} & \textbf{83.0} & 93.0 & \textbf{82.2} \\
\bottomrule
\end{tabular*}
\end{table}

Fig.~\ref{fig:supp_difficulty} breaks the results of Tab.~\ref{tab:paradigm} down by puzzle size.
The advantage of frame-wise causal generation grows with the size of the puzzle.
On Sokoban with three boxes, it solves 54.4\% of the test puzzles, compared with 11.8\% for bidirectional generation, and it remains the best on the unseen puzzles with four and five boxes.
On the quarter of Sokoban puzzles with the longest optimal solutions, it still solves 54.9\%, whereas bidirectional generation solves only 12.3\%.
On FlowFree, bidirectional generation fails on boards of every size, and chunk-wise generation stays below 35\% even on $5{\times}5$ boards.
CFT (Tab.~\ref{tab:posttrain}) further improves frame-wise generation on every puzzle size, and most on the harder puzzles, \eg, from 54.4\% to 74.9\% with three boxes.

\subsection{Causal Generation with Diffusion Forcing}
\label{supp:df}

\begin{table}[t]
\centering
\footnotesize
\setlength{\tabcolsep}{3pt}
\caption{\textbf{Teacher forcing and diffusion forcing} (SSR, \%). Both causal models generate with adaptive termination and are trained with the same data, schedule, and number of updates.}
\label{tab:supp_df}
\begin{tabular*}{\linewidth}{@{\hspace{\tabcolsep}\extracolsep{\fill}}lcccc}
\toprule
& \multicolumn{2}{c}{Wan2.2-5B} & \multicolumn{2}{c}{DiT-B} \\
\cmidrule(lr){2-3}\cmidrule(lr){4-5}
Method & FlowFree & Sokoban & FlowFree & Sokoban \\
\midrule
Bidirectional & 1.1 & 50.8 & 0.0 & 8.8 \\
HDR~\cite{hdr} & 11.6 & 45.9 & 0.4 & 10.3 \\
S2PD~\cite{s2pd} & 68.1 & 75.7 & 41.2 & 56.4 \\
\midrule
Ours, diffusion forcing & 69.9 & 81.8 & \textbf{59.6} & 58.1 \\
Ours, teacher forcing & \textbf{77.9} & \textbf{83.0} & 58.6 & \textbf{73.8} \\
\bottomrule
\end{tabular*}
\end{table}

Teacher forcing and diffusion forcing~\cite{diffusionforcing} are two ways to train a causal video model.
Teacher forcing conditions each frame on the clean history, whereas diffusion forcing trains every frame with an independent noise level, so that each frame is conditioned on a noisy history.
We train the frame-wise model of Tab.~\ref{tab:paradigm} with diffusion forcing, keeping its data, schedule, and number of updates, and follow its paper in re-noising the generated history to noise level $t{=}0.1$ at inference.
As Tab.~\ref{tab:supp_df} shows, causal generation with adaptive termination outperforms every fixed-length method whether it is trained with teacher forcing or diffusion forcing.
This indicates that the gain comes from generating the solution step by step and deciding when to stop, rather than from how the causal model is trained.
Teacher forcing is better in three of the four settings, by 6.0 points on average, and needs no re-noising at inference, so we use it in all other experiments.

\subsection{Sampling Steps and Shifts}
\label{supp:sampling}

\begin{table*}[t]
\centering
\footnotesize
\setlength{\tabcolsep}{4pt}
\caption{\textbf{Number of sampling steps} per segment (SSR, \%, and seconds per video), on stratified subsets of 120--200 test samples per benchmark. Ours is frame-wise generation, and chunk-wise generation on RoboTwin-Plan. HDR and S2PD use their default schedules, with 5--50 steps per level for HDR and 10 steps for S2PD. Time is measured on the same hardware at batch size 1.}
\label{tab:supp_steps}
\begin{tabular*}{\linewidth}{@{\hspace{\tabcolsep}\extracolsep{\fill}}lccccccccc}
\toprule
& & \multicolumn{2}{c}{FlowFree} & \multicolumn{2}{c}{Sokoban} & \multicolumn{2}{c}{VBVR-MultiStep} & \multicolumn{2}{c}{RoboTwin-Plan} \\
\cmidrule(lr){3-4}\cmidrule(lr){5-6}\cmidrule(lr){7-8}\cmidrule(lr){9-10}
Method & Steps & SSR & Time & SSR & Time & SSR & Time & SSR & Time \\
\midrule
\multirow{4}{*}{Bidirectional} & 2 & 0.0 & 2.5 & 28.5 & 2.5 & 10.8 & 2.3 & 11.7 & 2.2 \\
& 5 & 1.0 & 3.1 & 41.0 & 3.1 & 23.3 & 2.9 & 52.5 & 2.7 \\
& 10 & 1.5 & 4.1 & 42.5 & 4.0 & 25.8 & 3.8 & 65.8 & 3.5 \\
& 20 & 4.0 & 6.0 & 44.0 & 6.0 & 25.8 & 5.7 & 69.2 & 5.1 \\
\midrule
HDR & 5--50 & 9.0 & 188.2 & 34.0 & 184.2 & 27.5 & 180.4 & 67.5 & 140.8 \\  %
S2PD & 10 & 63.5 & 14.6 & 67.5 & 14.3 & 67.5 & 14.6 & 69.2 & 13.2 \\
\midrule
\multirow{4}{*}{Ours} & 2 & 60.0 & 5.8 & 73.5 & 7.7 & 79.2 & 4.3 & 55.8 & 3.3 \\
& 5 & \textbf{73.0} & 10.2 & \textbf{79.5} & 11.5 & 79.2 & 7.7 & \textbf{82.5} & 3.9 \\
& 10 & 71.5 & 16.6 & 79.0 & 18.8 & \textbf{81.7} & 12.7 & 80.8 & 4.9 \\
& 20 & 70.0 & 29.4 & 79.0 & 33.8 & 80.8 & 23.3 & 80.8 & 6.9 \\
\bottomrule
\end{tabular*}
\vspace{1mm}
\end{table*}

\begin{table}[tp]
\centering
\footnotesize
\setlength{\tabcolsep}{4pt}
\caption{\textbf{Training shift $s$ versus sampling shift} (SSR, \%, five steps), on the subsets of Tab.~\ref{tab:supp_steps}.}
\label{tab:supp_shift}
\begin{tabular*}{\linewidth}{@{\hspace{\tabcolsep}\extracolsep{\fill}}llccc}
\toprule
& & \multicolumn{3}{c}{Sampling shift} \\
\cmidrule(lr){3-5}
Benchmark & Training & 1 & 5 & 10 \\
\midrule
\multirow{3}{*}{FlowFree} & $s{=}1$ & 34.5 & 46.0 & 46.5 \\
& $s{=}5$ & 67.0 & 71.5 & 69.0 \\
& $s{=}10$ & \textbf{73.5} & \textbf{73.0} & \textbf{70.5} \\
\midrule
\multirow{3}{*}{Sokoban} & $s{=}1$ & 55.0 & 57.0 & 59.0 \\
& $s{=}5$ & 74.0 & 73.0 & 74.0 \\
& $s{=}10$ & \textbf{80.5} & \textbf{79.5} & \textbf{79.5} \\
\midrule
\multirow{3}{*}{VBVR-MultiStep} & $s{=}1$ & 40.8 & 42.5 & 42.5 \\
& $s{=}5$ & 71.7 & 73.3 & 72.5 \\
& $s{=}10$ & \textbf{82.5} & \textbf{79.2} & \textbf{79.2} \\
\midrule
\multirow{3}{*}{RoboTwin-Plan} & $s{=}1$ & 2.5 & 2.5 & 1.7 \\
& $s{=}5$ & 34.2 & 45.8 & 29.2 \\
& $s{=}10$ & \textbf{70.8} & \textbf{82.5} & \textbf{79.2} \\
\bottomrule
\end{tabular*}
\end{table}

\noindent\textbf{Sampling steps and inference cost.}
Tab.~\ref{tab:supp_steps} studies how the number of sampling steps per segment affects accuracy and inference time.
Our models change little beyond the five steps used in all other experiments, whereas bidirectional generation keeps improving but stays well below them, \eg, 69.2\% with 20 steps on RoboTwin-Plan against 82.5\% for our five.
With five steps, our models take 3.9--11.5\,s per video, compared with 2.7--3.1\,s for bidirectional generation, 141--188\,s for HDR, and 13.2--14.6\,s for S2PD, both of which are also less accurate.
At a similar time per video, they are far more accurate, \eg, 73.5\% with two steps against 44.0\% for bidirectional generation with 20 steps on Sokoban.

\noindent\textbf{Training and sampling shifts.}
The decision-focused schedule changes the shift used in training, while all models sample with a shift of 5.
Tab.~\ref{tab:supp_shift} evaluates the models of Tab.~\ref{tab:schedule} with sampling shifts of 1, 5, and 10.
The model trained with $s{=}10$ is the best under every sampling shift on all four benchmarks, so the gain of the decision-focused schedule does not depend on the sampler.
It also remains the best with only two sampling steps, \eg, 60.0\% against 49.5\% for $s{=}5$ on FlowFree.
On RoboTwin-Plan, the larger training shift also reduces visual defects of the arm and the blocks, which cause the first error in 55.6\%, 30.7\%, and 10.7\% of the test episodes with $s{=}1$, $5$, and $10$.
Most of these defects are distortions of the moving arm, whose new pose is generated at high noise levels, where the larger shift puts more training.

\subsection{Additional Post-Training Results}
\label{supp:posttrain_results}
\label{supp:vbvr_posttrain}

\label{supp:passk}

\noindent\textbf{pass@$k$.}
Tab.~\ref{tab:supp_passk} reports pass@$k$, computed from the same seeds for every model.
CFT achieves the highest pass@1 and pass@2 on all four game sets and the highest pass@4 on three of them.
Its gains also hold at $k{=}4$, \eg, 88.6\% on Sokoban-Hard compared with 82.2\% for continued SFT, so CFT raises the overall success rate rather than succeeding only with a particular seed.
On VBVR-MultiStep, CFT is also the best at every $k$.
On RoboTwin-Plan, CFT has the highest pass@1, while at $k{=}2$ and $k{=}4$ it is within 0.4 points of the best model, as 60\% of its failures there are visual defects of the arm or the blocks, about half of which a new sample avoids.

\noindent\textbf{Per-task results on VBVR-MultiStep.}
Tab.~\ref{tab:supp_vbvr_posttrain} lists the post-training results of Tab.~\ref{tab:posttrain} for each VBVR-MultiStep task.
CFT raises the SSR most on Tower of Hanoi, the task on which the SFT model is weakest, from 37.0\% to 61.0\%, and it also improves Snake Routing and Sokoban while keeping the nearly solved tasks at their level.
The policy-gradient methods leave Tower of Hanoi at or below the SFT model, and decision-level DiffusionNFT halves the SSR on Sliding Puzzle.

\subsection{Mechanism Analysis of CFT}
\label{supp:mechanism}

\begin{figure*}[t]
\centering
\includegraphics[width=0.85\linewidth]{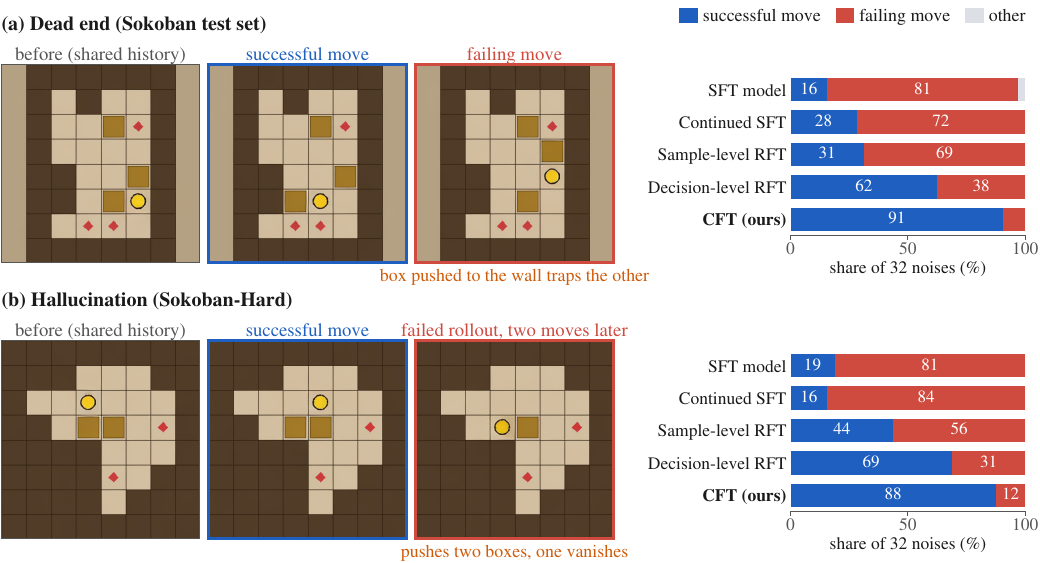}  %
\caption{\textbf{Two divergence points of the probe in Fig.~\ref{fig:mechanism}}, at which the failed rollout (a) reaches a dead end and (b) hallucinates. Left: the last frame of the shared history, the successful move, and the failing move, shown in (b) two moves later, where the failure becomes visible. Right: the share of 32 fresh noises with which each model generates the successful decision, the failing decision, or another move.}
\label{fig:supp_mechanism_example}
\end{figure*}

\begin{table}[tp]
\centering
\footnotesize
\setlength{\tabcolsep}{2.0pt}
\caption{\textbf{Decisions at the 149 divergence points} (\%). ``Fresh'' averages over 32 fresh noises per point, and ``failure noise'' uses the noise of the failed rollout. A move is safe if it is legal and the puzzle remains solvable after it.}
\label{tab:supp_mechanism}
\begin{tabular*}{\linewidth}{@{\hspace{\tabcolsep}\extracolsep{\fill}}lcccc}
\toprule
& \multicolumn{2}{c}{Successful decision} & \multicolumn{2}{c}{Safe move} \\
\cmidrule(lr){2-3}\cmidrule(lr){4-5}
Method & Fresh & Failure noise & Fresh & Failure noise \\
\midrule
SFT model (start) & 43.8 & 1.3 & 81.0 & 63.8 \\
+ Continued SFT & 49.3 & 20.1 & 84.2 & 73.8 \\
+ Sample-level RFT & 55.7 & 36.2 & 87.2 & 83.2 \\
+ Decision-level RFT & 65.9 & 48.3 & \textbf{95.0} & 90.6 \\
+ CFT (ours) & \textbf{67.9} & \textbf{57.0} & \textbf{95.0} & \textbf{93.3} \\
\bottomrule
\end{tabular*}
\end{table}

\noindent\textbf{Probe construction.}
The probe uses the models of Tab.~\ref{tab:posttrain} on Sokoban.
We roll out the SFT model on the test sets and keep every failed rollout for which a continuation resampled from an earlier step succeeds, exactly as CFT collects its training data, which gives 67 points on the test set and 82 on Sokoban-Hard.
At each point, every model generates the next decision from the shared history $\mathbf{h}_d$, once with the noise $\boldsymbol{\epsilon}^-_d$ of the failed rollout and once with each of 32 fresh noises, which are the same for all models.
The verifier parses the board after the decision and labels it as the successful decision, the failing decision, or another move.
Fig.~\ref{fig:supp_mechanism_example} shows two examples, at which CFT generates the successful decision with 91\% and 88\% of the fresh noises, compared with 16\% and 19\% for the SFT model.

\noindent\textbf{Successful decisions and safe moves.}
Tab.~\ref{tab:supp_mechanism} complements Fig.~\ref{fig:mechanism}.
We also count safe moves, \ie, legal moves after which the puzzle remains solvable, which we check with an exhaustive Sokoban solver.
The ranking of the methods is the same, but this metric separates them less.
The failed rollout often loses the puzzle only one to four steps after the divergence, so its decision at the divergence point is itself safe at 63.8\% of the points, although it leads to failure.
We therefore measure the successful decision in the main paper.

\subsection{Ablations and Extensions of CFT}
\label{supp:cft_variants}
\label{supp:qt}

\begin{table}[tp]
\centering
\footnotesize
\setlength{\tabcolsep}{3pt}
\caption{\textbf{Noise mapping and timestep distribution} on FlowFree (SSR, \%). Fresh noise with $p_{\mathrm{DF}}(t)$ is decision-level RFT, and the noise of the failed rollout with $q(t)$ is CFT.}
\label{tab:supp_noise_qt}
\begin{tabular*}{\linewidth}{@{\hspace{\tabcolsep}\extracolsep{\fill}}llcccc}
\toprule
& & \multicolumn{2}{c}{Wan2.2-5B} & \multicolumn{2}{c}{DiT-B} \\
\cmidrule(lr){3-4}\cmidrule(lr){5-6}
Noise & Timesteps & Test & Hard & Test & Hard \\
\midrule
Fresh & $p_{\mathrm{DF}}(t)$ & 80.1 & 52.0 & 59.0 & \textbf{28.0} \\
Fresh & $q(t)$ & 80.8 & 48.6 & 62.1 & 27.0 \\
Failed rollout & $p_{\mathrm{DF}}(t)$ & 83.6 & \textbf{53.6} & 63.4 & 25.4 \\
Failed rollout & $q(t)$ & \textbf{84.4} & 52.0 & \textbf{65.6} & 27.2 \\
\bottomrule
\end{tabular*}
\end{table}

\noindent\textbf{Noise mapping and timestep distribution.}
CFT differs from decision-level RFT in two ways: it reuses the noise of the failed rollout, and its corrective loss places extra samples at $t{=}1$ (Eq.~\eqref{eq:cft_corr}).
Tab.~\ref{tab:supp_noise_qt} separates the two on FlowFree.
Reusing the failure noise brings the larger gain on the test set under either timestep distribution, 3.5--3.6 points on Wan2.2-5B and 3.5--4.4 points on DiT-B, compared with 0.7--0.8 and 2.2--3.1 points for the extra samples at $t{=}1$.
Combining both, as CFT does, gives the best test SSR on both backbones.

\begin{table}[tp]
\centering
\footnotesize
\setlength{\tabcolsep}{3pt}
\caption{\textbf{Corrective timestep distribution $q(t)$} on DiT-B (SSR, \%). $P(t{=}1)$ is the probability of placing a corrective sample at $t{=}1$; the remaining samples follow $p_{\mathrm{DF}}(t)$. CFT uses $0.5$.}
\label{tab:supp_qt}
\begin{tabular*}{\linewidth}{@{\hspace{\tabcolsep}\extracolsep{\fill}}lcccccc}
\toprule
& \multicolumn{2}{c}{$P(t{=}1){=}0$} & \multicolumn{2}{c}{$P(t{=}1){=}0.5$} & \multicolumn{2}{c}{$P(t{=}1){=}1$} \\
\cmidrule(lr){2-3}\cmidrule(lr){4-5}\cmidrule(lr){6-7}
Game & Test & Hard & Test & Hard & Test & Hard \\
\midrule
FlowFree & 63.4 & 25.4 & \textbf{65.6} & \textbf{27.2} & 61.3 & 23.6 \\
Sokoban & 79.1 & 46.1 & \textbf{79.6} & \textbf{48.3} & 79.2 & 47.9 \\
\bottomrule
\end{tabular*}
\end{table}

\noindent\textbf{Corrective timestep distribution.}
CFT samples the timestep of the corrective loss (Eq.~\eqref{eq:cft_corr}) from $q(t)$, which places a corrective sample at $t{=}1$ with probability $P(t{=}1)$ and otherwise draws $t$ from $p_{\mathrm{DF}}(t)$.
At $t{=}1$, the noisy input is exactly the noise $\boldsymbol{\epsilon}^-_d$ of the failed rollout, so the model learns to map that noise to the successful decision, while the other timesteps train the denoising path toward it.
As Tab.~\ref{tab:supp_qt} shows, the mixed choice $P(t{=}1){=}0.5$ is best on both games and both test sets, and training only at $t{=}1$ or never at $t{=}1$ is worse.

\label{supp:multiround}
\looseness=-1
\noindent\textbf{Multi-round CFT.}
Tab.~\ref{tab:cft_analysis} shows that a second round of CFT improves the first one further, and we give the details here.
In the second round, we roll out the first-round CFT model on the training puzzles, including the 5,000 harder ones, and collect new corrections with the procedure of the first round.
This gives 5,201 successful continuations from 1,601 puzzles on FlowFree and 3,758 on Sokoban.
Both the second round and the control start from the first-round model and use the same optimizer, learning-rate schedule, and number of updates, and the control trains on the first-round data instead.

\label{supp:solver}

\begin{figure}[t]
\centering
\includegraphics[width=0.87\linewidth]{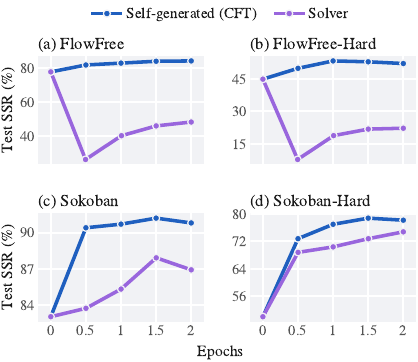}  %
\caption{\textbf{Self-generated versus solver corrections} on Wan2.2-5B (test SSR, \%), evaluated every half epoch. Both start from the same SFT model at epoch 0 and train for two epochs over their own corrections.}
\label{fig:supp_solver}
\end{figure}

\noindent\textbf{Self-generated versus solver corrections.}
For rule-based games like FlowFree and Sokoban, an exact solver can provide a correction for every failed rollout, including those for which no resampled continuation succeeds.
We test whether such corrections can replace the self-generated ones of CFT.
Starting from the same branch points as CFT, we complete each failed rollout with the solver, render the completion with our renderer, and encode the video.
On Sokoban, we use a shortest solution, and on FlowFree, the solution that keeps the path of the failed rollout for as long as the puzzle remains solvable.
Training uses the same objective as CFT, which pairs the solver segment at the divergence step with the noise of the failed rollout, as well as the same expert share and learning-rate schedule, and runs for two epochs over its own data (11,324 and 9,388 updates with batch 2).
This gives 8,494 corrections from 6,259 puzzles on FlowFree and 7,039 from 4,601 puzzles on Sokoban, 2.8 and 1.7 times as many puzzles as the self-generated corrections (Tab.~\ref{tab:supp_posttrain}).

As Fig.~\ref{fig:supp_solver} shows, solver corrections are worse than self-generated ones at every checkpoint on both games, although they cover more puzzles.
On FlowFree, they fall far below the SFT model, reaching 48.2\% on the test set and 22.2\% on FlowFree-Hard after two epochs, compared with 84.4\% and 52.0\% for CFT, as the model often stops before the board is filled or breaks a path.
On Sokoban, they improve over the SFT model but stay below CFT, with 86.9\% and 74.8\% compared with 90.8\% and 78.2\%.
Self-generated corrections lie within the model's own distribution and only change which decision the noise leads to.
By contrast, solver corrections lie outside it, and fitting them disturbs the generation itself.
Corrections are therefore most effective when drawn from the model's own distribution, even though a solver covers more puzzles.

\subsection{Out-of-Distribution Generalization}
\label{supp:ood}

\begin{figure*}[t]
\centering
\includegraphics[width=\linewidth]{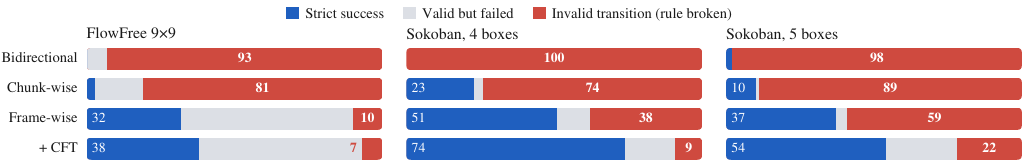}  %
\caption{\textbf{Failure analysis on the larger puzzles} (\%, 100 puzzles each), with each video categorized by its first error as in Fig.~\ref{fig:rule_breaks}. CFT starts from the frame-wise model.}
\label{fig:supp_ood_failures}
\end{figure*}

\noindent\textbf{Comparison of post-training methods.}
Tab.~\ref{tab:supp_ood_posttrain} evaluates the post-trained models of Tab.~\ref{tab:posttrain} on the larger puzzles of Sec.~\ref{sec:exp_ood}.
CFT is the best method on all three sets, tied with continued SFT on FlowFree.
The policy-gradient methods mostly fall below the SFT model on these puzzles.

\begin{table}[tbp]
\centering
\footnotesize
\setlength{\tabcolsep}{3pt}
\caption{\textbf{Post-training methods on the larger puzzles} (SSR, \%, 100 puzzles each).}
\label{tab:supp_ood_posttrain}
\begin{tabular*}{\linewidth}{@{\hspace{\tabcolsep}\extracolsep{\fill}}lccc}
\toprule
& FlowFree & \multicolumn{2}{c}{Sokoban} \\
\cmidrule(lr){2-2}\cmidrule(lr){3-4}
Method & 9$\times$9 & 4 boxes & 5 boxes \\
\midrule
SFT model (ours) & 32.0 & 51.0 & 37.0 \\
+ Continued SFT & \textbf{38.0} & 58.0 & 33.0 \\
\catrow{4}{Sample-level}
+ Flow-GRPO & 30.0 & 45.0 & 22.0 \\
+ DiffusionNFT & 33.0 & 45.0 & 28.0 \\
+ RFT & 30.0 & 56.0 & 39.0 \\
\catrow{4}{Decision-level}
+ Flow-GRPO & 31.0 & 38.0 & 26.0 \\
+ DiffusionNFT & 18.0 & 36.0 & 26.0 \\
+ RFT & 31.0 & 65.0 & 45.0 \\
+ CFT (ours) & \textbf{38.0} & \textbf{74.0} & \textbf{54.0} \\
\bottomrule
\end{tabular*}
\end{table}

\noindent\textbf{Failure analysis.}
Fig.~\ref{fig:supp_ood_failures} categorizes the videos on the larger puzzles by their first error.
On FlowFree, frame-wise generation breaks a rule in only 10\% of the videos, compared with 93\% for bidirectional generation, and most of its failures follow every rule but leave the board unfinished.
On Sokoban, rule violations remain the main failure of frame-wise generation as the number of boxes grows, and CFT more than halves them, from 38\% to 9\% of the videos with four boxes and from 59\% to 22\% with five.

\noindent\textbf{New layouts and longer plans.}
Tab.~\ref{tab:supp_ood_robotwin} lists all out-of-distribution sets of RoboTwin-Plan.
Chunk-wise generation, the best paradigm on this benchmark, generalizes best to the new layouts and to longer plans, and bidirectional generation solves none of the longer plans.
CFT, which starts from the chunk-wise model, breaks the rules about as often on these sets and leaves fewer plans unfinished, but its videos show more visual defects of the arm and the blocks, which lowers its success rate (Tab.~\ref{tab:supp_rt_errors}). The same effect appears on the original test set, where every method that fine-tunes on its own rollouts shows more visual defects than the start model, whereas continued SFT on rendered expert videos shows fewer. There, CFT's large reduction of rule violations and unfinished plans outweighs these defects.

\begin{table}[tbp]
\centering
\footnotesize
\setlength{\tabcolsep}{2.5pt}
\caption{\textbf{First errors on RoboTwin-Plan} (\% of episodes) of the chunk-wise start model and the post-training methods of Tab.~\ref{tab:posttrain}. Rule: a task rule is broken; unfinished: the video ends before the goal; visual: a visual defect of the arm or the blocks. New sets: the 150 episodes of Tab.~\ref{tab:supp_ood_robotwin}.}
\label{tab:supp_rt_errors}
\begin{tabular*}{\linewidth}{@{\hspace{\tabcolsep}\extracolsep{\fill}}llcccc}
\toprule
Test set & Method & Success & Rule & Unfinished & Visual \\
\midrule
\multirow{5}{*}{Original} & Ours ($K{=}8$) & 73.9 & 7.7 & 7.7 & 10.7 \\
& + Continued SFT & 74.3 & 8.9 & 8.6 & 8.3 \\
& + Sample-level RFT & 77.9 & 7.0 & 3.1 & 11.9 \\
& + Decision-level RFT & 70.7 & 6.1 & 5.4 & 17.7 \\
& + CFT (ours) & 78.0 & 4.9 & 4.1 & 13.0 \\
\midrule
\multirow{2}{*}{New sets} & Ours ($K{=}8$) & 26.7 & 38.0 & 14.7 & 20.7 \\
& + CFT (ours) & 21.3 & 38.0 & 8.7 & 32.0 \\
\bottomrule
\end{tabular*}
\end{table}

\begin{table}[tbp]
\centering
\footnotesize
\setlength{\tabcolsep}{2.6pt}
\caption{\textbf{RoboTwin-Plan beyond the training distribution} (SSR, \%). The first two sets apply trained rules to new layouts; the last one requires longer plans. CFT starts from the chunk-wise model.}
\label{tab:supp_ood_robotwin}
\begin{tabular*}{\linewidth}{@{\hspace{\tabcolsep}\extracolsep{\fill}}lccc}
\toprule
& Blocks World & Hanoi & Longer \\
Method & 2$\times$2 (50) & 2$\times$2 (50) & plans (50) \\  %
\midrule
Bidirectional & 14.0 & 16.0 & 0.0 \\
Ours ($K{=}8$) & \textbf{36.0} & \textbf{18.0} & \textbf{26.0} \\
Ours ($K{=}1$) & 4.0 & 2.0 & 22.0 \\
+ CFT (ours) & 32.0 & 14.0 & 18.0 \\
\bottomrule
\end{tabular*}
\end{table}

\begin{table}[t]
\centering
\footnotesize
\setlength{\tabcolsep}{3pt}
\caption{\textbf{Comparison with frontier multimodal language models} (SSR, \%) on 50 random puzzles of each game test set. Language models answer with moves as text, which the same verifiers replay.}
\label{tab:supp_llm}
\begin{tabular*}{\linewidth}{@{\hspace{\tabcolsep}\extracolsep{\fill}}lcccc}
\toprule
& \multicolumn{2}{c}{FlowFree} & \multicolumn{2}{c}{Sokoban} \\
\cmidrule(lr){2-3}\cmidrule(lr){4-5}
Method & Test & Hard & Test & Hard \\
\midrule
Qwen3.8-27B & 68.0 & 42.0 & 36.0 & 40.0 \\
Claude Opus 5.5 & \textbf{100.0} & \textbf{100.0} & \textbf{100.0} & \textbf{96.0} \\
\ours{} (video) & 82.0 & 66.0 & 90.0 & 82.0 \\
\bottomrule
\end{tabular*}
\end{table}

\begin{figure*}[t]  %
\centering
\includegraphics[width=0.98\linewidth]{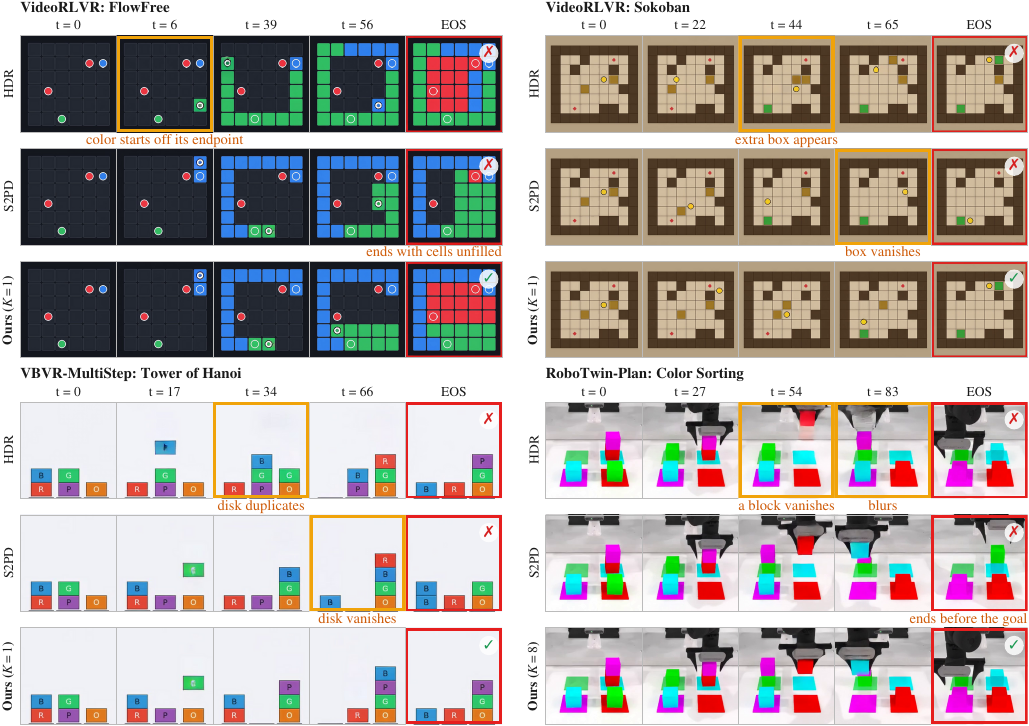}  %
\caption{\textbf{Comparison with HDR and S2PD} on one test puzzle per benchmark. Orange borders mark rule violations.}
\label{fig:supp_baselines}
\end{figure*}

\subsection{Comparison with Frontier Multimodal Language Models}
\label{supp:llm}

\looseness=-1  %
Given the rapid progress of LLMs, a natural question is how video models compare with the frontier ones, which can reason about such puzzles in text.
We give two of them, Claude Opus 5.5\footnote{\url{https://www.anthropic.com/claude}} and the open Qwen3.8-27B\footnote{\url{https://huggingface.co/Qwen/Qwen3.8-27B}}, the first frame and the task prompt of each puzzle and ask for the moves as text.
The same verifiers replay their answers, and Tab.~\ref{tab:supp_llm} compares them with \ours{} on 50 random puzzles from each game test set.
Qwen3.8-27B runs in thinking mode with its default sampling settings and up to 32K new tokens.
Opus 5.5 solves almost every puzzle, while \ours{} outperforms Qwen3.8-27B by 14 to 54 points.
The two settings differ in what the model must produce.
A language model answers with a short move sequence after a long reasoning trace, a median of 11.2K tokens for Qwen3.8-27B, whereas a video model solves the puzzle by generating every intermediate state as a frame, so a single wrong frame fails the video.
The video also models how the scene changes physically, \eg, a robot arm grasping, carrying, and placing blocks on RoboTwin-Plan, which a text answer leaves out.
As an academic study, we use open models to explore how far video models can reason by generating their own state transitions, and combining them with language-model planners is a promising direction.

\renewcommand{\dbltopfraction}{0.9}\renewcommand{\textfraction}{0.07}\renewcommand{\dblfloatpagefraction}{0.85}
\section{Additional Qualitative Results}
\label{supp:qualitative}

\begin{figure*}[tp]
\centering
\includegraphics[width=0.985\linewidth]{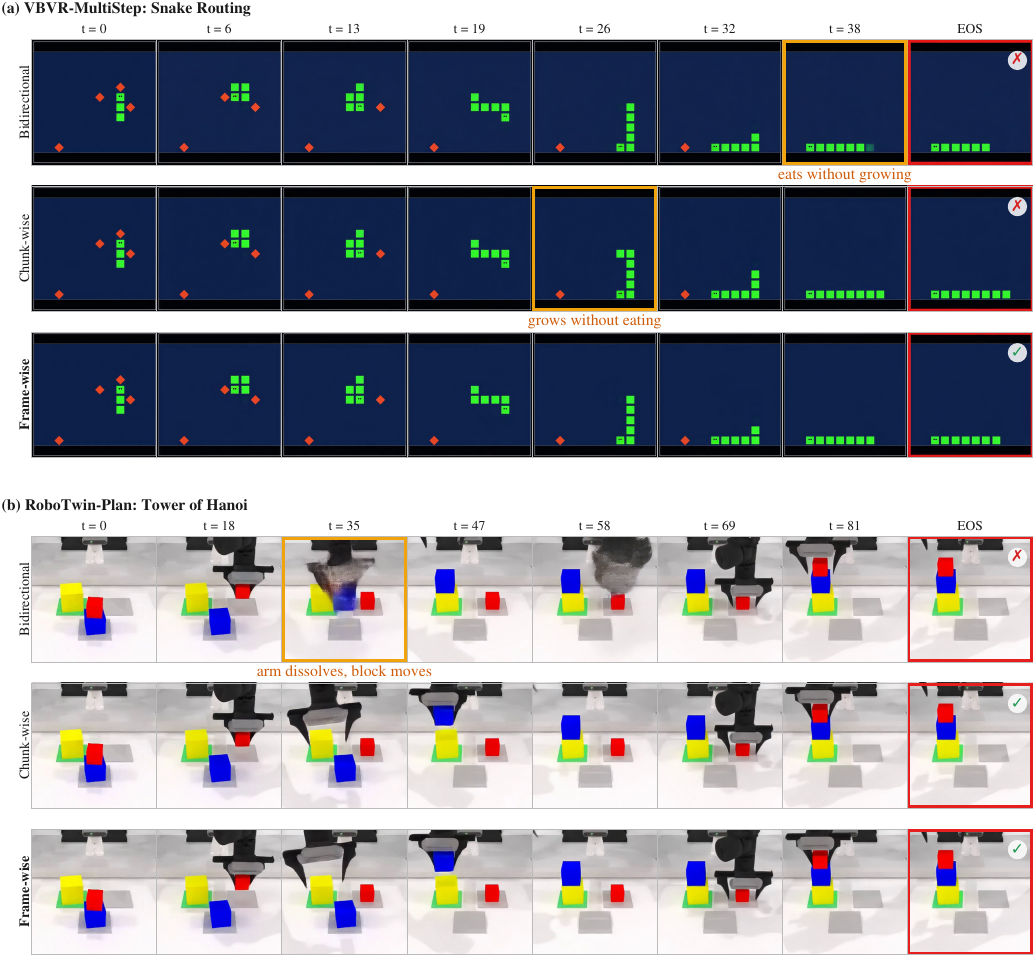}  %
\caption{\textbf{More comparisons of the generation paradigms} on two tasks not included in Fig.~\ref{fig:qualitative}. Orange borders mark rule violations, and the last column shows the final frame of each video.}
\label{fig:supp_paradigms}
\end{figure*}

\begin{figure*}[tp]
\centering
\includegraphics[width=\linewidth]{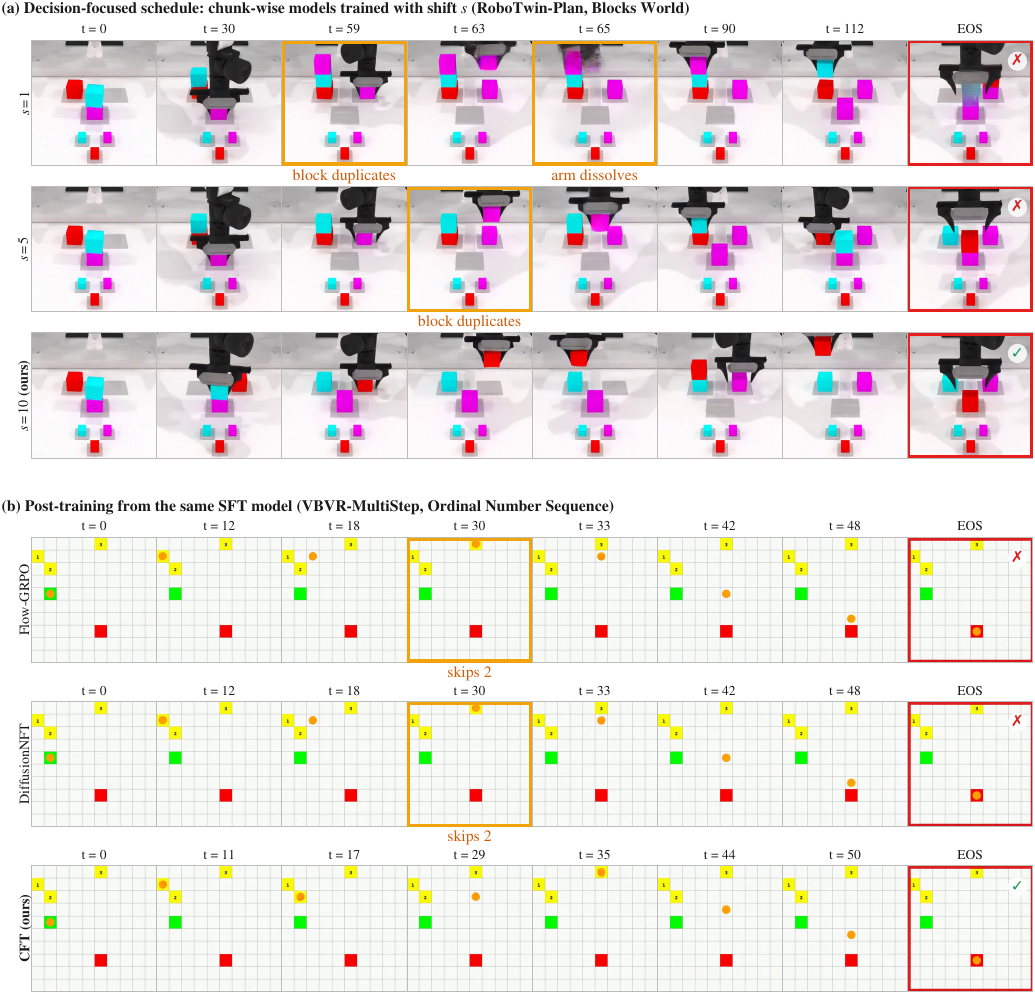}  %
\caption{\textbf{Qualitative results of (a) the decision-focused schedule} on RoboTwin-Plan \textbf{and (b) post-training} on VBVR-MultiStep. Orange borders mark rule violations, and the last column shows the final frame of each video.}
\label{fig:supp_qualitative}
\end{figure*}

\begin{figure*}[tp]
\centering
\includegraphics[width=\linewidth]{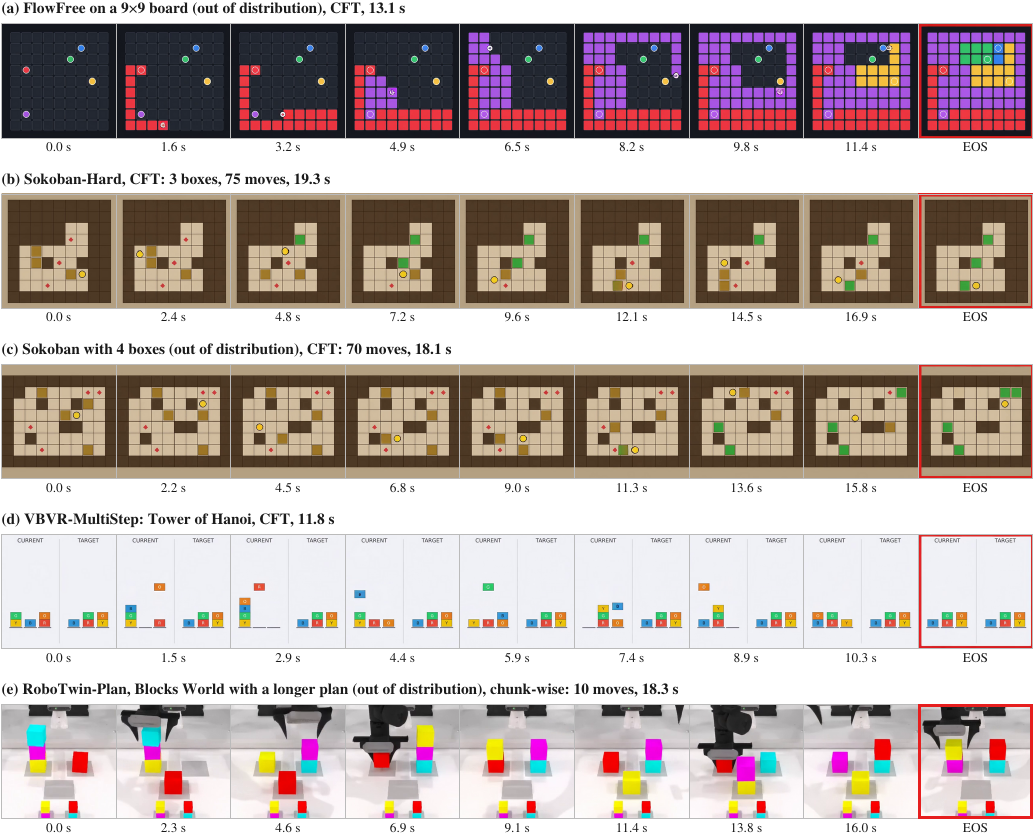}  %
\caption{\textbf{Long-horizon solutions.} The longest strictly successful videos of our models, with frames evenly spaced from the first frame to the EOS frame. The longest expert videos last 15.8\,s on Sokoban, 11.6\,s on Blocks World, 10.8\,s on FlowFree, and 12.1\,s on Tower of Hanoi.}
\label{fig:supp_longhorizon}
\end{figure*}

We first compare our models with the prior methods HDR and S2PD in Fig.~\ref{fig:supp_baselines}, using one test puzzle per benchmark. The examples show common failures of the two baselines, such as boxes that appear or vanish in Sokoban and plans that end before the goal on RoboTwin-Plan.

We also compare the generation paradigms in Fig.~\ref{fig:supp_paradigms} on two tasks not included in Fig.~\ref{fig:qualitative}, Snake Routing and Tower of Hanoi of RoboTwin-Plan. Frame-wise generation solves both, whereas bidirectional generation breaks the rules on both.

Fig.~\ref{fig:supp_qualitative} shows how the decision-focused schedule and CFT improve the generated videos. With $s{=}1$ and $s{=}5$, the chunk-wise model duplicates a block midway, and after sample-level Flow-GRPO and DiffusionNFT, the model skips a number, whereas $s{=}10$ and CFT follow the rules.

Finally, Fig.~\ref{fig:supp_longhorizon} shows the longest strictly successful videos of our models. CFT solves a Sokoban-Hard puzzle in 75 moves within 19.3\,s, and chunk-wise generation completes a 10-move Blocks World plan in 18.3\,s, whereas the expert videos of these tasks last at most 15.8\,s and 11.6\,s.

\section{Limitations and Future Work}
\label{supp:limitations}

\noindent\textbf{Scope of training.}
To keep the comparisons clean and controlled within our compute budget, we train one model per benchmark on its own training data.
Training a single model on larger and more diverse reasoning data, possibly mixed with general video data, is a natural next step toward video models that solve a broad range of tasks.

\noindent\textbf{Rule verifiers.}
Like other methods based on verifiable rewards~\cite{videorlvr,vbvrpro}, CFT relies on a verifier, which here checks every state transition to locate the failures that CFT corrects.
Such verifiers can be built agentically for tasks with explicit rules (Sec.~\ref{supp:evaluation}), but are harder to obtain for open-ended tasks.
Learned verifiers or VLM judges could extend CFT to these tasks.

\noindent\textbf{From videos to actions.}
Our work focuses on improving video models themselves, and RoboTwin-Plan evaluates the generated plans as videos in simulation.
Executing them on real robots requires converting the videos into actions, \eg, with an inverse dynamics model~\cite{unipi}, which we leave to future work.
As recent world-action models build on video generation models~\cite{cosmospolicy,dreamzero,vera,dyna2}, video models with stronger physical consistency and longer-horizon planning could also provide a better foundation for them.

\clearpage
\begin{table*}[p]  %
\centering
\footnotesize  %
\setlength{\tabcolsep}{4pt}
\caption{\textbf{Task prompts} used in training and testing. Angle brackets mark parts that change with the puzzle, \eg, the colors of the maze theme, the number of moves of a Sliding Puzzle, or the number and shape of the TSP rewards. The two layouts of each RoboTwin-Plan task share one prompt.}
\label{tab:supp_prompts}
\begin{tabular}{@{}llp{0.66\linewidth}@{}}
\toprule
Benchmark & Task & Prompt \\
\midrule
\multirow{8}{*}{VideoRLVR} & Maze & Solve the maze by drawing one continuous $\langle$color$\rangle$ path from the $\langle$color$\rangle$ start to the $\langle$color$\rangle$ goal without crossing a wall. \\
 & FlowFree & Fill the grid one cell at a time. Start at any colored endpoint and visit every remaining endpoint of that color. Then switch at an endpoint of a new color. Moves are horizontal or vertical, cannot cross or overlap, and every cell must be filled. Keep a single continuous walk across color changes: the next endpoint must be adjacent to the previous filled cell. \\
 & Sokoban & Solve the Sokoban puzzle. Move the agent and push every box onto a target. Boxes cannot be pulled and walls cannot be crossed. \\
\midrule
\multirow{35}{*}{VBVR-MultiStep} & Tower of Hanoi & [Scenario] The image shows colored blocks arranged in stacks. There is a CURRENT state on the left and a TARGET state on the right. [Rules] 1. Only the topmost block of any stack can be moved at a time. 2. A move consists of lifting a block, moving it horizontally, and lowering it onto another stack or empty position. [Task] Generate a video solving the puzzle by transforming the CURRENT state to exactly match the TARGET state. Animate the blocks moving one by one, using the minimum total number of moves required. \\
 & Ordinal Number Sequence & [Scenario] The image shows a grid with a green start point, a red end point, and several yellow cells marked with sequential numbers (1, 2, 3...). An orange circular agent is positioned at the start. [Rules] 1. The agent can only move orthogonally (up, down, left, right). 2. The agent MUST visit the numbered yellow cells in strictly increasing numerical order (1, then 2, then 3, etc.). 3. The agent must reach the red end point only after all numbered cells have been visited in order. [Task] Generate a video tracking the orange agent as it visits the numbers sequentially and heads to the final goal. The animation must reflect the shortest valid path for each sequential segment. \\
 & Sliding Puzzle & [Scenario] The image displays a sliding tile puzzle grid. The tiles are numbered, and there is exactly one blank space. [Rules] 1. Only tiles orthogonally adjacent to the blank space can be moved. 2. Only slide one tile per move into the blank space. 3. The goal state requires tiles to be arranged in sequential order (top to bottom, left to right), with the blank space at the bottom-right corner. [Task] Generate a video showing the completion of this sliding puzzle. Animate the exact sequence of $\langle$n$\rangle$ moves required to reach the goal state without any extra moves. Keep the camera view fixed. \\
 & Snake Routing & [Scenario] The image shows a snake game board with a snake and multiple red food pellets scattered across the grid. [Rules] 1. The snake must eat all the food pellets on the board. 2. Each pellet eaten grows the snake's length by exactly one cell. 3. The snake strictly cannot collide with its own body or the grid boundaries. [Task] Generate a video animating the snake moving to consume every food pellet. The snake must follow the shortest possible total path to collect all items without violating any rules. \\
 & Sokoban & [Scenario] The image shows a classic Sokoban puzzle. The red character is the player, blue crates are the boxes, yellow glowing circles are target positions, and brown brick blocks are walls. [Rules] 1. The player can move up, down, left, or right. 2. The player can push only one box at a time into an empty space. Boxes cannot be pulled. 3. Walls cannot be moved through or pushed. [Task] Generate a video solving the Sokoban puzzle. Animate the player moving and pushing the boxes step-by-step until all boxes are placed onto the target positions. \\
 & TSP Reward Collection & [Scenario] The image displays a 10x10 grid with a green start point, a red end point, and $\langle$k$\rangle$ $\langle$shape$\rangle$ reward items scattered across it. A circular agent begins at the start point. [Rules] 1. The agent can only move to adjacent cells (up, down, left, right). 2. The agent collects a reward simply by moving onto its cell; collected rewards disappear. 3. All $\langle$k$\rangle$ $\langle$shape$\rangle$ rewards must be collected BEFORE reaching the red end point. [Task] Generate a video animating the circular agent collecting all rewards and finishing at the red end point. Trace the agent's path as it moves along the absolute shortest valid route. \\
\midrule
\multirow{8}{*}{RoboTwin-Plan} & Blocks World & Two robot arms move one top block at a time between the three grey places on the table until the stacks match the small model at the front of the table. \\
 & Tower of Hanoi & Two robot arms move all blocks onto the green pad, one top block at a time, never placing a larger block on a smaller one. \\
 & Color Sorting (buffer) & Two robot arms move one top block at a time until every block stands alone on the pad of its own colour; the grey pad is a free buffer. \\
 & Color Sorting (no buffer) & Two robot arms move one top block at a time until every block stands alone on the pad of its own colour; there is no free pad, so a blocking block can only be put on top of another block. \\
\bottomrule
\end{tabular}
\end{table*}

%% file: main.bbl
\begin{thebibliography}{64}
\providecommand{\natexlab}[1]{#1}
\providecommand{\url}[1]{\texttt{#1}}
\expandafter\ifx\csname urlstyle\endcsname\relax
  \providecommand{\doi}[1]{doi: #1}\else
  \providecommand{\doi}{doi: \begingroup \urlstyle{rm}\Url}\fi

\bibitem[Bergmeister et~al.(2026)Bergmeister, Jegelka, N{\"u}sken,
  Domingo-Enrich, and Pidstrigach]{ram}
Andreas Bergmeister, Stefanie Jegelka, Nikolas N{\"u}sken, Carles
  Domingo-Enrich, and Jakiw Pidstrigach.
\newblock {Reinforce Adjoint Matching: Scaling RL Post-Training of Diffusion
  and Flow Models}.
\newblock In \emph{Advances in Neural Information Processing Systems}, 2026.

\bibitem[Black et~al.(2024)Black, Janner, Du, Kostrikov, and Levine]{ddpo}
Kevin Black, Michael Janner, Yilun Du, Ilya Kostrikov, and Sergey Levine.
\newblock {Training Diffusion Models with Reinforcement Learning}.
\newblock In \emph{International Conference on Learning Representations
  (ICLR)}, 2024.

\bibitem[{ByteDance Seed}(2026)]{seedance25}
{ByteDance Seed}.
\newblock {One-take Creation, Flexible Referencing: Introducing Seedance 2.5}.
\newblock Blog post, 2026.

\bibitem[Chen et~al.(2024)Chen, Mons{\'o}, Du, Simchowitz, Tedrake, and
  Sitzmann]{diffusionforcing}
Boyuan Chen, Diego~Mart{\'i} Mons{\'o}, Yilun Du, Max Simchowitz, Russ Tedrake,
  and Vincent Sitzmann.
\newblock {Diffusion Forcing: Next-token Prediction Meets Full-Sequence
  Diffusion}.
\newblock In \emph{Advances in Neural Information Processing Systems}, pages
  24081--24125, 2024.

\bibitem[Chen et~al.(2026{\natexlab{a}})Chen, Chen, Chen, Cai, Liu, Li, Liang,
  Lin, Ge, Gu, Deng, Guo, Nian, Xie, Chen, Su, Xu, Liu, Hu, Gao, Wang, Liang,
  Qin, Yang, Luo, and Mu]{robotwin}
Tianxing Chen, Zanxin Chen, Baijun Chen, Zijian Cai, Yibin Liu, Zixuan Li,
  Qiwei Liang, Xianliang Lin, Yiheng Ge, Zhenyu Gu, Weiliang Deng, Yubin Guo,
  Tian Nian, Xuanbing Xie, Qiangyu Chen, Kailun Su, Tianling Xu, Guodong Liu,
  Mengkang Hu, {Huan-ang} Gao, Kaixuan Wang, Zhixuan Liang, Yusen Qin, Xiaokang
  Yang, Ping Luo, and Yao Mu.
\newblock {RoboTwin 2.0: A Scalable Data Generator and Benchmark with Strong
  Domain Randomization for Robust Bimanual Robotic Manipulation}.
\newblock In \emph{Proceedings of the 43rd International Conference on Machine
  Learning (ICML)}, pages 13673--13699. PMLR, 2026{\natexlab{a}}.

\bibitem[Chen et~al.(2026{\natexlab{b}})Chen, Zhang, Guo, Jiang, and
  Li]{opencof}
Xinyan Chen, Renrui Zhang, Ziyu Guo, Dongzhi Jiang, and Hongsheng Li.
\newblock {OpenCoF: Learning to Reason Through Video Generation}.
\newblock In \emph{Advances in Neural Information Processing Systems},
  2026{\natexlab{b}}.

\bibitem[Cheng et~al.(2026)Cheng, Hou, Zhong, Tao, Wan, Gai, and
  Liao]{vlmasteacher}
Junhao Cheng, Liang Hou, Tianxiong Zhong, Xin Tao, Pengfei Wan, Kun Gai, and
  Jing Liao.
\newblock {VLMs are Good Teachers for Video Reasoning via Adaptive Test-Time
  Optimization}.
\newblock \emph{arXiv preprint arXiv:2606.02564}, 2026.

\bibitem[Choi et~al.(2026)Choi, Guo, Zhu, Vahdat, Tao, Berner, and Chen]{rvm}
Jaemoo Choi, Wei Guo, Yuchen Zhu, Arash Vahdat, Molei Tao, Julius Berner, and
  Yongxin Chen.
\newblock {Scaling Reinforcement Learning for Diffusion Models via Velocity
  Matching}.
\newblock \emph{arXiv preprint arXiv:2608.23664}, 2026.

\bibitem[Diaz~Chao et~al.(2026)Diaz~Chao, Preechakul, Liu, and
  Bai]{serialitygap}
Jorge Diaz~Chao, Konpat Preechakul, Yuxi Liu, and Yutong Bai.
\newblock The seriality gap in video diffusion models.
\newblock \emph{arXiv preprint arXiv:2607.13031}, 2026.

\bibitem[Dong et~al.(2023)Dong, Xiong, Goyal, Zhang, Chow, Pan, Diao, Zhang,
  Shum, and Zhang]{raft}
Hanze Dong, Wei Xiong, Deepanshu Goyal, Yihan Zhang, Winnie Chow, Rui Pan,
  Shizhe Diao, Jipeng Zhang, Kashun Shum, and Tong Zhang.
\newblock {RAFT: Reward rAnked FineTuning for Generative Foundation Model
  Alignment}.
\newblock \emph{Transactions on Machine Learning Research}, 2023.

\bibitem[Du et~al.(2023)Du, Yang, Dai, Dai, Nachum, Tenenbaum, Schuurmans, and
  Abbeel]{unipi}
Yilun Du, Sherry Yang, Bo Dai, Hanjun Dai, Ofir Nachum, Joshua~B. Tenenbaum,
  Dale Schuurmans, and Pieter Abbeel.
\newblock {Learning Universal Policies via Text-Guided Video Generation}.
\newblock In \emph{Advances in Neural Information Processing Systems}, pages
  9156--9172, 2023.

\bibitem[Du et~al.(2024)Du, Yang, Florence, Xia, Wahid, Ichter, Sermanet, Yu,
  Abbeel, Tenenbaum, Kaelbling, Zeng, and Tompson]{vlp}
Yilun Du, Sherry Yang, Pete Florence, Fei Xia, Ayzaan Wahid, Brian Ichter,
  Pierre Sermanet, Tianhe Yu, Pieter Abbeel, Joshua~B. Tenenbaum, Leslie
  Kaelbling, Andy Zeng, and Jonathan Tompson.
\newblock {Video Language Planning}.
\newblock In \emph{International Conference on Learning Representations
  (ICLR)}, 2024.

\bibitem[{Dyna Robotics}(2026)]{dyna2}
{Dyna Robotics}.
\newblock {Dyna-2: A 1-Million-Hour Scaling Law for World-Action Models}.
\newblock Technical report, 2026.

\bibitem[Geirhos et~al.(2026)Geirhos, Li, Wiedemer, Kalibhat, Wang, Malek,
  Tafjord, Swersky, Kim, and Jaini]{visualpromptengineering}
Robert Geirhos, Yuxuan Li, Thadd{\"a}us Wiedemer, Neha Kalibhat, Zi Wang, Mani
  Malek, Oyvind Tafjord, Kevin Swersky, Been Kim, and Priyank Jaini.
\newblock {Visual prompt engineering for video models}.
\newblock \emph{arXiv preprint arXiv:2607.25537}, 2026.

\bibitem[{Google DeepMind}(2025)]{veo31}
{Google DeepMind}.
\newblock {Veo 3.1}.
\newblock Model documentation, Vertex AI (\texttt{veo-3.1-generate-001}), 2025.

\bibitem[{Google DeepMind}(2026)]{genie4}
{Google DeepMind}.
\newblock {Genie 4: A New Generation of World Simulation}.
\newblock Blog post, 2026.

\bibitem[Guo et~al.(2026{\natexlab{a}})Guo, Wu, Wang, Mireshghallah, and
  Kavraki]{scalingvideoreasoning}
Weihang Guo, Xiaoyu Wu, Yifei Wang, Niloofar Mireshghallah, and Lydia~E.
  Kavraki.
\newblock {Scaling Video Generation for Reasoning: At What Cost?}
\newblock \emph{arXiv preprint arXiv:2609.36599}, 2026{\natexlab{a}}.

\bibitem[Guo et~al.(2026{\natexlab{b}})Guo, Chen, Zhang, An, Qi, Jiang, Li,
  Zhang, Li, and Heng]{mme_cof}
Ziyu Guo, Xinyan Chen, Renrui Zhang, Ruichuan An, Yu Qi, Dongzhi Jiang,
  Xiangtai Li, Manyuan Zhang, Hongsheng Li, and Pheng-Ann Heng.
\newblock {Are Video Models Ready as Zero-Shot Reasoners? An Empirical Study
  with the MME-CoF Benchmark}.
\newblock In \emph{Proceedings of the IEEE/CVF Conference on Computer Vision
  and Pattern Recognition (CVPR) Findings}, pages 9175--9184,
  2026{\natexlab{b}}.

\bibitem[He et~al.(2026{\natexlab{a}})He, Feng, Ge, Zhang, Ma, Song, Liu, and
  Li]{arcopo}
Dailan He, Guanlin Feng, Xingtong Ge, Yi Zhang, Bingqi Ma, Guanglu Song, Yu
  Liu, and Hongsheng Li.
\newblock {AR-CoPO: Align Autoregressive Video Generation with Contrastive
  Policy Optimization}.
\newblock In \emph{European Conference on Computer Vision (ECCV)}, pages
  57--73, 2026{\natexlab{a}}.

\bibitem[He et~al.(2026{\natexlab{b}})He, Wei, Cheng, Ma, Zhang, Li, Wen,
  Jiang, Liu, Hao, Cai, Wu, Du, Zou, Yang, Sun, Yang, Nie, Allen, Wang, Galley,
  Gao, and Zhai]{vgibench}
Xuan He, Cong Wei, Yuhao Cheng, Linrui Ma, Yuxuan Zhang, Zuojun Li, Yuhao Wen,
  Jize Jiang, Zeyi Liu, Yuren Hao, Songcheng Cai, Keming Wu, Penghui Du, Kai
  Zou, Rui Yang, Chenkai Sun, Ke Yang, Ping Nie, Kelsey~R. Allen, Chenglong
  Wang, Michel Galley, Jianfeng Gao, and ChengXiang Zhai.
\newblock {VGI-Bench: Probing Visual Intelligence in Video Generation Models}.
\newblock \emph{arXiv preprint arXiv:2608.19583}, 2026{\natexlab{b}}.

\bibitem[Hu et~al.(2026)Hu, Reino, and Tewari]{s2pd}
Jeffrey Hu, Daniel~Olmeda Reino, and Ayush Tewari.
\newblock {S2PD: Serial-to-Parallel Diffusion for Physically and Logically
  Consistent Video Generation}.
\newblock \emph{arXiv preprint arXiv:2610.06847}, 2026.

\bibitem[Huang et~al.(2025)Huang, Li, He, Zhou, and Shechtman]{selfforcing}
Xun Huang, Zhengqi Li, Guande He, Mingyuan Zhou, and Eli Shechtman.
\newblock {Self Forcing: Bridging the Train-Test Gap in Autoregressive Video
  Diffusion}.
\newblock In \emph{Advances in Neural Information Processing Systems}, pages
  167283--167308, 2025.

\bibitem[Huang et~al.(2026)Huang, Yu, Chen, Qiu, Debevec, and Liu]{vchain}
Ziqi Huang, Ning Yu, Gordon Chen, Haonan Qiu, Paul Debevec, and Ziwei Liu.
\newblock {VChain: Chain-of-Visual-Thought for Reasoning in Video Generation}.
\newblock In \emph{Findings of the Association for Computational Linguistics:
  ACL 2026}, pages 226--250, 2026.

\bibitem[Jun et~al.(2026)Jun, Ding, Su, Ye, and Du]{temporalbacktracking}
SeJoon Jun, Zheng Ding, Huangyuan Su, Weirui Ye, and Yilun Du.
\newblock {Temporal Backtracking Search for Test-time Generative Video
  Reasoning}.
\newblock In \emph{Advances in Neural Information Processing Systems}, 2026.

\bibitem[Kim et~al.(2026)Kim, Gao, Lin, Lin, Ge, Lam, Liang, Song, Liu, Finn,
  and Gu]{cosmospolicy}
Moo~Jin Kim, Yihuai Gao, Tsung-Yi Lin, Yen-Chen Lin, Yunhao Ge, Grace Lam,
  Percy Liang, Shuran Song, Ming-Yu Liu, Chelsea Finn, and Jinwei Gu.
\newblock {Cosmos Policy: Fine-Tuning Video Models for Visuomotor Control and
  Planning}.
\newblock In \emph{International Conference on Learning Representations
  (ICLR)}, 2026.

\bibitem[Li et~al.(2026{\natexlab{a}})Li, Kim, Bai, Zhao, Pang, Simchowitz, and
  Sitzmann]{vera}
Sizhe~Lester Li, Evan Kim, Xingjian Bai, Tong Zhao, Tao Pang, Max Simchowitz,
  and Vincent Sitzmann.
\newblock {Turning Video Models into Generalist Robot Policies}.
\newblock \emph{arXiv preprint arXiv:2605.27817}, 2026{\natexlab{a}}.

\bibitem[Li et~al.(2026{\natexlab{b}})Li, Gu, Min, Liu, Du, Zhou, Yang, Zhao,
  and Qiu]{viper}
Yifan Li, YuKai Gu, Yingqian Min, Zikang Liu, Yifan Du, Kun Zhou, Min Yang, Xin
  Zhao, and Minghui Qiu.
\newblock {Beyond the Last Frame: Process-aware Evaluation for Generative Video
  Reasoning}.
\newblock In \emph{Proceedings of the 64th Annual Meeting of the Association
  for Computational Linguistics (Volume 1: Long Papers)}, pages 20393--20409,
  2026{\natexlab{b}}.

\bibitem[Lipman et~al.(2023)Lipman, Chen, Ben-Hamu, Nickel, and
  Le]{flowmatching}
Yaron Lipman, Ricky T.~Q. Chen, Heli Ben-Hamu, Maximilian Nickel, and Matthew
  Le.
\newblock Flow matching for generative modeling.
\newblock In \emph{International Conference on Learning Representations}, 2023.

\bibitem[Liu et~al.(2025)Liu, Liu, Liang, Li, Liu, Wang, Wan, Zhang, and
  Ouyang]{flowgrpo}
Jie Liu, Gongye Liu, Jiajun Liang, Yangguang Li, Jiaheng Liu, Xintao Wang,
  Pengfei Wan, Di Zhang, and Wanli Ouyang.
\newblock {Flow-GRPO: Training Flow Matching Models via Online RL}.
\newblock In \emph{Advances in Neural Information Processing Systems}, pages
  40783--40818, 2025.

\bibitem[Liu et~al.(2026{\natexlab{a}})Liu, Zhang, Liu, Wang, and Zhang]{wanr1}
Ming Liu, Yunbei Zhang, Shilong Liu, Liwen Wang, and Wensheng Zhang.
\newblock {Wan-R1: Verifiable-Reinforcement Learning for Video Reasoning}.
\newblock In \emph{European Conference on Computer Vision (ECCV)}, pages
  419--436, 2026{\natexlab{a}}.

\bibitem[Liu et~al.(2023)Liu, Gong, and Liu]{rectifiedflow}
Xingchao Liu, Chengyue Gong, and Qiang Liu.
\newblock Flow straight and fast: Learning to generate and transfer data with
  rectified flow.
\newblock In \emph{International Conference on Learning Representations}, 2023.

\bibitem[Liu et~al.(2026{\natexlab{b}})Liu, Preechakul, Kuwaranancharoen, and
  Bai]{serialscaling}
Yuxi Liu, Konpat Preechakul, Kananart Kuwaranancharoen, and Yutong Bai.
\newblock The serial scaling hypothesis.
\newblock In \emph{International Conference on Learning Representations},
  2026{\natexlab{b}}.

\bibitem[Ma et~al.(2024)Ma, Liang, Wang, Huang, Bastani, Jayaraman, Zhu, Fan,
  and Anandkumar]{eureka}
Yecheng~Jason Ma, William Liang, Guanzhi Wang, De-An Huang, Osbert Bastani,
  Dinesh Jayaraman, Yuke Zhu, Linxi Fan, and Anima Anandkumar.
\newblock {Eureka: Human-Level Reward Design via Coding Large Language Models}.
\newblock In \emph{International Conference on Learning Representations
  (ICLR)}, 2024.

\bibitem[Newman et~al.(2026)Newman, Zhu, and
  Russakovsky]{videomodelsreasonearly}
Kaleb Newman, Tyler Zhu, and Olga Russakovsky.
\newblock {Video Models Reason Early: Exploiting Plan Commitment for Maze
  Solving}.
\newblock \emph{arXiv preprint arXiv:2603.30043}, 2026.

\bibitem[Ni et~al.(2026)Ni, Li, Jiao, Zhan, Chen, Yin, Chen, Torr, Wang, and
  Zhu]{creflow}
Zhenyang Ni, Yijiang Li, Ruochen Jiao, Simon~Sinong Zhan, Sipeng Chen, Zhenfei
  Yin, Minshuo Chen, Philip Torr, Zhaoran Wang, and Qi Zhu.
\newblock {CreFlow: Corrective Reflow for Sparse-Reward Embodied Video
  Diffusion RL}.
\newblock \emph{arXiv preprint arXiv:2605.14274}, 2026.

\bibitem[{NVIDIA}(2026)]{cosmos3}
{NVIDIA}.
\newblock {Cosmos 3: Omnimodal World Models for Physical AI}.
\newblock \emph{arXiv preprint arXiv:2606.02800}, 2026.

\bibitem[Peebles and Xie(2023)]{dit}
William Peebles and Saining Xie.
\newblock {Scalable Diffusion Models with Transformers}.
\newblock In \emph{Proceedings of the IEEE/CVF International Conference on
  Computer Vision (ICCV)}, pages 4195--4205, 2023.

\bibitem[Qian et~al.(2026)Qian, Chi, Mak, Zhou, Yuan, Rui, Sun, Wu, Li, Qian,
  Han, and Zhang]{hdr}
Zezhong Qian, Xiaowei Chi, Chak-Wing Mak, Tianze Zhou, Ruibin Yuan, Yuhan Rui,
  Hengzhe Sun, Zhuoqun Wu, Yuming Li, Siyuan Qian, Sirui Han, and Shanghang
  Zhang.
\newblock {Hierarchical Denoising For Multi-Step Visual Reasoning}.
\newblock In \emph{Advances in Neural Information Processing Systems}, 2026.

\bibitem[Ren et~al.(2025)Ren, Wei, Guo, Zhao, Kang, Feng, and Jin]{videoworld}
Zhongwei Ren, Yunchao Wei, Xun Guo, Yao Zhao, Bingyi Kang, Jiashi Feng, and
  Xiaojie Jin.
\newblock {VideoWorld: Exploring Knowledge Learning from Unlabeled Videos}.
\newblock In \emph{Proceedings of the IEEE/CVF Conference on Computer Vision
  and Pattern Recognition (CVPR)}, pages 29029--29039, 2025.

\bibitem[Ren et~al.(2026)Ren, Wei, Yu, Luo, Zhao, Kang, Feng, and
  Jin]{videoworld2}
Zhongwei Ren, Yunchao Wei, Xiao Yu, Guixun Luo, Yao Zhao, Bingyi Kang, Jiashi
  Feng, and Xiaojie Jin.
\newblock {VideoWorld 2: Learning Transferable Knowledge from Real-world
  Videos}.
\newblock In \emph{Proceedings of the IEEE/CVF Conference on Computer Vision
  and Pattern Recognition (CVPR)}, pages 40569--40580, 2026.

\bibitem[Schulman et~al.(2017)Schulman, Wolski, Dhariwal, Radford, and
  Klimov]{ppo}
John Schulman, Filip Wolski, Prafulla Dhariwal, Alec Radford, and Oleg Klimov.
\newblock {Proximal Policy Optimization Algorithms}.
\newblock \emph{arXiv preprint arXiv:1707.06347}, 2017.

\bibitem[Shao et~al.(2024)Shao, Wang, Zhu, Xu, Song, Bi, Zhang, Zhang, Li, Wu,
  and Guo]{grpo}
Zhihong Shao, Peiyi Wang, Qihao Zhu, Runxin Xu, Junxiao Song, Xiao Bi, Haowei
  Zhang, Mingchuan Zhang, Y.~K. Li, Y. Wu, and Daya Guo.
\newblock {DeepSeekMath: Pushing the Limits of Mathematical Reasoning in Open
  Language Models}.
\newblock \emph{arXiv preprint arXiv:2402.03300}, 2024.

\bibitem[Shen et~al.(2026)Shen, Zhu, Zhang, and Chen]{proar}
Linghui Shen, Tinghui Zhu, Sheng Zhang, and Muhao Chen.
\newblock {ProAR: Learning Prospective Reasoning with Autoregressive Video
  Models}.
\newblock \emph{arXiv preprint arXiv:2610.03664}, 2026.

\bibitem[Sim{\'e}oni et~al.(2026)Sim{\'e}oni, Vo, Seitzer, Baldassarre, Oquab,
  Jose, Khalidov, Szafraniec, Yi, Ramamonjisoa, Massa, Haziza, Wehrstedt, Wang,
  Darcet, Moutakanni, Sentana, Roberts, Vedaldi, Tolan, Brandt, Couprie,
  Mairal, J{\'e}gou, Labatut, and Bojanowski]{dinov3}
Oriane Sim{\'e}oni, Huy~V. Vo, Maximilian Seitzer, Federico Baldassarre, Maxime
  Oquab, Cijo Jose, Vasil Khalidov, Marc Szafraniec, Seungeun Yi, Micha{\"e}l
  Ramamonjisoa, Francisco Massa, Daniel Haziza, Luca Wehrstedt, Jianyuan Wang,
  Timoth{\'e}e Darcet, Th{\'e}o Moutakanni, Leonel Sentana, Claire Roberts,
  Andrea Vedaldi, Jamie Tolan, John Brandt, Camille Couprie, Julien Mairal,
  Herv{\'e} J{\'e}gou, Patrick Labatut, and Piotr Bojanowski.
\newblock {DINOv3}.
\newblock \emph{Transactions on Machine Learning Research}, 2026.

\bibitem[{Team Wan} et~al.(2025){Team Wan}, Wang, Ai, Wen, Mao, Xie, Chen, Yu,
  Zhao, Yang, Zeng, Wang, Zhang, Zhou, Wang, Chen, Zhu, Zhao, Yan, Huang, Feng,
  Zhang, Li, Wu, Chu, Feng, Zhang, Sun, Fang, Wang, Gui, Weng, Shen, Lin, Wang,
  Wang, Zhou, Wang, Shen, Yu, Shi, Huang, Xu, Kou, Lv, Li, Liu, Wang, Zhang,
  Huang, Li, Wu, Liu, Pan, Zheng, Hong, Shi, Feng, Jiang, Han, Wu, and
  Liu]{wan}
{Team Wan}, Ang Wang, Baole Ai, Bin Wen, Chaojie Mao, Chen-Wei Xie, Di Chen,
  Feiwu Yu, Haiming Zhao, Jianxiao Yang, Jianyuan Zeng, Jiayu Wang, Jingfeng
  Zhang, Jingren Zhou, Jinkai Wang, Jixuan Chen, Kai Zhu, Kang Zhao, Keyu Yan,
  Lianghua Huang, Mengyang Feng, Ningyi Zhang, Pandeng Li, Pingyu Wu, Ruihang
  Chu, Ruili Feng, Shiwei Zhang, Siyang Sun, Tao Fang, Tianxing Wang, Tianyi
  Gui, Tingyu Weng, Tong Shen, Wei Lin, Wei Wang, Wei Wang, Wenmeng Zhou, Wente
  Wang, Wenting Shen, Wenyuan Yu, Xianzhong Shi, Xiaoming Huang, Xin Xu, Yan
  Kou, Yangyu Lv, Yifei Li, Yijing Liu, Yiming Wang, Yingya Zhang, Yitong
  Huang, Yong Li, You Wu, Yu Liu, Yulin Pan, Yun Zheng, Yuntao Hong, Yupeng
  Shi, Yutong Feng, Zeyinzi Jiang, Zhen Han, Zhi-Fan Wu, and Ziyu Liu.
\newblock {Wan: Open and Advanced Large-Scale Video Generative Models}.
\newblock \emph{arXiv preprint arXiv:2503.20314}, 2025.

\bibitem[Tong et~al.(2026)Tong, Mou, Li, Li, Yang, Zhang, Chen, Liang, Hu,
  Zheng, Chen, Zhao, Huang, and Qiu]{thinkingwithvideo}
Jingqi Tong, Yurong Mou, Hangcheng Li, Mingzhe Li, Yongzhuo Yang, Ming Zhang,
  Qiguang Chen, Tianyi Liang, Xiaomeng Hu, Yining Zheng, Xinchi Chen, Jun Zhao,
  Xuanjing Huang, and Xipeng Qiu.
\newblock {Thinking with Video: Video Generation as a Promising Multimodal
  Reasoning Paradigm}.
\newblock In \emph{Proceedings of the IEEE/CVF Conference on Computer Vision
  and Pattern Recognition (CVPR)}, pages 41121--41129, 2026.

\bibitem[Wan et~al.(2026)Wan, Chen, Zhang, Zhou, Zhao, Kong, Wan, Gai, and
  You]{diffusionart}
Zhenglin Wan, Jiayu Chen, Zhengfeiyang Zhang, Pengfei Zhou, Wangbo Zhao,
  Fangyuan Kong, Pengfei Wan, Kun Gai, and Yang You.
\newblock {DiffusionART: Efficient, Stable and Scalable Reinforcement of
  Diffusion Models}.
\newblock \emph{arXiv preprint arXiv:2610.12572}, 2026.

\bibitem[Wang et~al.(2026{\natexlab{a}})Wang, Wang, Lin, Ji, Wiedemer, Gao,
  Luo, Qian, Huang, Hong, Ge, Ma, He, Zhou, Guo, Mei, Li, Xing, Zhao, Yu, Xiao,
  Jiao, Hou, Zhang, Xu, Zhong, Zhao, Fang, Kitaoka, Xu, Xu, Blacutt, Nguyen,
  Song, Sun, Wen, He, Wang, Wang, Yang, Ma, Milli{\`e}re, Shi, Vasconcelos,
  Khashabi, Yuille, Du, Liu, Lin, Liu, Kumar, Li, Yang, Cai, and Deng]{vbvr}
Maijunxian Wang, Ruisi Wang, Juyi Lin, Ran Ji, Thadd{\"a}us Wiedemer, Qingying
  Gao, Dezhi Luo, Yaoyao Qian, Lianyu Huang, Zelong Hong, Jiahui Ge, Qianli Ma,
  Hang He, Yifan Zhou, Lingzi Guo, Lantao Mei, Jiachen Li, Hanwen Xing, Tianqi
  Zhao, Fengyuan Yu, Weihang Xiao, Yizheng Jiao, Jianheng Hou, Danyang Zhang,
  Pengcheng Xu, Boyang Zhong, Zehong Zhao, Gaoyun Fang, John Kitaoka, Yile Xu,
  Hua Xu, Kenton Blacutt, Tin Nguyen, Siyuan Song, Haoran Sun, Shaoyue Wen,
  Linyang He, Runming Wang, Yanzhi Wang, Mengyue Yang, Ziqiao Ma, Rapha{\"e}l
  Milli{\`e}re, Freda Shi, Nuno Vasconcelos, Daniel Khashabi, Alan Yuille,
  Yilun Du, Ziming Liu, Dahua Lin, Ziwei Liu, Vikash Kumar, Yijiang Li, Lei
  Yang, Zhongang Cai, and Hokin Deng.
\newblock {A Very Big Video Reasoning Suite}.
\newblock In \emph{Proceedings of the 43rd International Conference on Machine
  Learning (ICML)}. PMLR, 2026{\natexlab{a}}.

\bibitem[Wang et~al.(2026{\natexlab{b}})Wang, Cai, Pu, Xu, Yin, Wang, Ji, Gu,
  Li, Huang, Deng, Lin, Liu, and Yang]{demystifyingvideoreasoning}
Ruisi Wang, Zhongang Cai, Fanyi Pu, Junxiang Xu, Wanqi Yin, Maijunxian Wang,
  Ran Ji, Chenyang Gu, Bo Li, Ziqi Huang, Hokin Deng, Dahua Lin, Ziwei Liu, and
  Lei Yang.
\newblock {Demystifying Video Reasoning}.
\newblock In \emph{European Conference on Computer Vision (ECCV)}, pages
  308--328, 2026{\natexlab{b}}.

\bibitem[Wang et~al.(2026{\natexlab{c}})Wang, Wang, Zhang, Zuo, Wu, Wang, Sun,
  Wang, Cao, Zhao, Guo, and Zhao]{worldcompass}
Zehan Wang, Tengfei Wang, Haiyu Zhang, Xuhui Zuo, Junta Wu, Haoyuan Wang,
  Wenqiang Sun, Zhenwei Wang, Chenjie Cao, Hengshuang Zhao, Chunchao Guo, and
  Zhou Zhao.
\newblock {WorldCompass: Reinforcement Learning for Long-Horizon World Models}.
\newblock In \emph{Proceedings of the 43rd International Conference on Machine
  Learning (ICML)}, pages 131140--131155. PMLR, 2026{\natexlab{c}}.

\bibitem[Wiedemer et~al.(2025)Wiedemer, Li, Vicol, Gu, Matarese, Swersky, Kim,
  Jaini, and Geirhos]{videomodelszeroshot}
Thadd{\"a}us Wiedemer, Yuxuan Li, Paul Vicol, Shixiang~Shane Gu, Nick Matarese,
  Kevin Swersky, Been Kim, Priyank Jaini, and Robert Geirhos.
\newblock {Video models are zero-shot learners and reasoners}.
\newblock \emph{arXiv preprint arXiv:2509.20328}, 2025.

\bibitem[Xie et~al.(2024)Xie, Zhao, Wu, Liu, Luo, Zhong, Yang, and
  Yu]{text2reward}
Tianbao Xie, Siheng Zhao, Chen~Henry Wu, Yitao Liu, Qian Luo, Victor Zhong,
  Yanchao Yang, and Tao Yu.
\newblock {Text2Reward: Reward Shaping with Language Models for Reinforcement
  Learning}.
\newblock In \emph{International Conference on Learning Representations
  (ICLR)}, 2024.

\bibitem[Xu et~al.(2026)Xu, Wang, Pu, Wang, Ji, Zhou, Gu, Zuo, Xiao, Geng, Yin,
  Chen, Qian, Yan, Huang, Diao, Pan, Li, Fan, Luo, Yu, Zhao, Gao, Zhu, Zhang,
  Tong, Feng, Jiang, Wang, Guo, Zhang, Chen, Joseph, Venhoff, Motamed, Yang,
  Sripada, Yuille, Torr, Zhang, Kumar, Khashabi, Kriegeskorte, Milli{\`e}re,
  M{\"u}ller, Rao, Wang, Liu, Lin, Yang, Deng, and Cai]{vbvrpro}
Junxiang Xu, Ruisi Wang, Fanyi Pu, Maijunxian Wang, Ran Ji, Tongxi Zhou,
  Chenyang Gu, Jing Zuo, Hongcan Xiao, Yimeng Geng, Wanqi Yin, Wei Chen, Oscar
  Qian, Zhengan Yan, Ziqi Huang, Haiwen Diao, Liang Pan, Bo Li, Xiangyu Fan,
  Dezhi Luo, Fengyuan Yu, Zehong Zhao, Qingying Gao, Tinghui Zhu, Yilan Zhang,
  Jingqi Tong, Pinyuan Feng, Zhengze Jiang, Letian Wang, Ziyu Guo, Renrui
  Zhang, Jieneng Chen, Sonia Joseph, Constantin Venhoff, Saman Motamed, Mengyue
  Yang, Chandra Sripada, Alan Yuille, Philip Torr, Lvmin Zhang, Vikash Kumar,
  Daniel Khashabi, Nikolaus Kriegeskorte, Rapha{\"e}l Milli{\`e}re, Vincent~C.
  M{\"u}ller, Anyi Rao, Quan Wang, Ziwei Liu, Dahua Lin, Lei Yang, Hokin Deng,
  and Zhongang Cai.
\newblock {VBVR-Pro: A Scalable and Verifiable Suite for Native Visual
  Reasoning}.
\newblock \emph{arXiv preprint arXiv:2608.26105}, 2026.

\bibitem[Xue et~al.(2026)Xue, Ge, Zhang, Li, and Ma]{awm}
Shuchen Xue, Chongjian Ge, Shilong Zhang, Yichen Li, and Zhi-Ming Ma.
\newblock {Advantage Weighted Matching: Aligning RL with Pretraining in
  Diffusion Models}.
\newblock In \emph{Proceedings of the 43rd International Conference on Machine
  Learning (ICML)}, pages 143146--143169. PMLR, 2026.

\bibitem[Xue et~al.(2025)Xue, Wu, Gao, Kong, Zhu, Chen, Liu, Liu, Guo, Huang,
  and Luo]{dancegrpo}
Zeyue Xue, Jie Wu, Yu Gao, Fangyuan Kong, Lingting Zhu, Mengzhao Chen, Zhiheng
  Liu, Wei Liu, Qiushan Guo, Weilin Huang, and Ping Luo.
\newblock {DanceGRPO: Unleashing GRPO on Visual Generation}.
\newblock \emph{arXiv preprint arXiv:2505.07818}, 2025.

\bibitem[Yang et~al.(2025)Yang, Wan, Peng, Cheng, Yu, Zhang, Yu, Yu, Zheng,
  Zhou, and Wu]{vrbench}
Cheng Yang, Haiyuan Wan, Yiran Peng, Xin Cheng, Zhaoyang Yu, Jiayi Zhang,
  Junchi Yu, Xinlei Yu, Xiawu Zheng, Dongzhan Zhou, and Chenglin Wu.
\newblock {Reasoning via Video: The First Evaluation of Video Models' Reasoning
  Abilities through Maze-Solving Tasks}.
\newblock \emph{arXiv preprint arXiv:2511.15065}, 2025.

\bibitem[Yang et~al.(2024)Yang, Du, Ghasemipour, Tompson, Kaelbling,
  Schuurmans, and Abbeel]{unisim}
Sherry Yang, Yilun Du, Kamyar Ghasemipour, Jonathan Tompson, Leslie Kaelbling,
  Dale Schuurmans, and Pieter Abbeel.
\newblock {Learning Interactive Real-World Simulators}.
\newblock In \emph{International Conference on Learning Representations
  (ICLR)}, 2024.

\bibitem[Ye et~al.(2026)Ye, Ge, Zheng, Gao, Yu, Kurian, Indupuru, Tan, Zhu,
  Xiang, Malik, Lee, Liang, Ranawaka, Gu, Xu, Wang, Hu, Narayan, Bjorck, Wang,
  Kim, Niu, Zheng, Xie, Wu, Wang, Julian, Xu, Du, Chebotar, Reed, Kautz, Zhu,
  Fan, and Jang]{dreamzero}
Seonghyeon Ye, Yunhao Ge, Kaiyuan Zheng, Shenyuan Gao, Sihyun Yu, George
  Kurian, Suneel Indupuru, You~Liang Tan, Chuning Zhu, Jiannan Xiang, Ayaan
  Malik, Kyungmin Lee, William Liang, Nadun Ranawaka, Jiasheng Gu, Yinzhen Xu,
  Guanzhi Wang, Fengyuan Hu, Avnish Narayan, Johan Bjorck, Jing Wang, Gwanghyun
  Kim, Dantong Niu, Ruijie Zheng, Yuqi Xie, Jimmy Wu, Qi Wang, Ryan Julian,
  Danfei Xu, Yilun Du, Yevgen Chebotar, Scott Reed, Jan Kautz, Yuke Zhu,
  Linxi~Jim Fan, and Joel Jang.
\newblock {World Action Models are Zero-shot Policies}.
\newblock In \emph{Proceedings of The 10th Conference on Robot Learning}, 2026.

\bibitem[Yin et~al.(2025)Yin, Zhang, Zhang, Freeman, Durand, Shechtman, and
  Huang]{causvid}
Tianwei Yin, Qiang Zhang, Richard Zhang, William~T. Freeman, Fr{\'e}do Durand,
  Eli Shechtman, and Xun Huang.
\newblock {From Slow Bidirectional to Fast Autoregressive Video Diffusion
  Models}.
\newblock In \emph{Proceedings of the IEEE/CVF Conference on Computer Vision
  and Pattern Recognition (CVPR)}, pages 22963--22974, 2025.

\bibitem[Yuan et~al.(2023)Yuan, Yuan, Li, Dong, Lu, Tan, Zhou, and Zhou]{rft}
Zheng Yuan, Hongyi Yuan, Chengpeng Li, Guanting Dong, Keming Lu, Chuanqi Tan,
  Chang Zhou, and Jingren Zhou.
\newblock {Scaling Relationship on Learning Mathematical Reasoning with Large
  Language Models}.
\newblock \emph{arXiv preprint arXiv:2308.01825}, 2023.

\bibitem[Zhang et~al.(2025)Zhang, Tang, Hu, Pan, Guo, Liu, Huang, Yuan, Zhang,
  Long, Cao, and Yin]{epona}
Kaiwen Zhang, Zhenyu Tang, Xiaotao Hu, Xingang Pan, Xiaoyang Guo, Yuan Liu,
  Jingwei Huang, Li Yuan, Qian Zhang, Xiao-Xiao Long, Xun Cao, and Wei Yin.
\newblock {Epona: Autoregressive Diffusion World Model for Autonomous Driving}.
\newblock In \emph{Proceedings of the IEEE/CVF International Conference on
  Computer Vision (ICCV)}, pages 27220--27230, 2025.

\bibitem[Zheng et~al.(2026)Zheng, Chen, Ye, Wang, Zhang, Jiang, Su, Ermon, Zhu,
  and Liu]{diffusionnft}
Kaiwen Zheng, Huayu Chen, Haotian Ye, Haoxiang Wang, Qinsheng Zhang, Kai Jiang,
  Hang Su, Stefano Ermon, Jun Zhu, and Ming-Yu Liu.
\newblock {DiffusionNFT: Online Diffusion Reinforcement with Forward Process}.
\newblock In \emph{International Conference on Learning Representations
  (ICLR)}, 2026.

\bibitem[Zhu et~al.(2026{\natexlab{a}})Zhu, Zhao, He, Su, Li, and
  Zhu]{causalforcing}
Hongzhou Zhu, Min Zhao, Guande He, Hang Su, Chongxuan Li, and Jun Zhu.
\newblock {Causal Forcing: Autoregressive Diffusion Distillation Done Right for
  High-Quality Real-Time Interactive Video Generation}.
\newblock In \emph{Proceedings of the 43rd International Conference on Machine
  Learning (ICML)}, pages 168030--168049. PMLR, 2026{\natexlab{a}}.

\bibitem[Zhu et~al.(2026{\natexlab{b}})Zhu, Zhang, Huang, Song, Wen, Li, Poon,
  and Chen]{videorlvr}
Tinghui Zhu, Sheng Zhang, James~Y. Huang, Selena Song, Xiaofei Wen, Yuankai Li,
  Hoifung Poon, and Muhao Chen.
\newblock {Video Models Can Reason with Verifiable Rewards}.
\newblock In \emph{Advances in Neural Information Processing Systems},
  2026{\natexlab{b}}.

\end{thebibliography}
